%% file: fax.tex
\documentclass{article}

\usepackage[preprint, nonatbib]{neurips_2020}

\usepackage[utf8]{inputenc} % allow utf-8 input
\usepackage[T1]{fontenc}    % use 8-bit T1 fonts
\usepackage{url}            % simple URL typesetting
\usepackage{booktabs ,csvsimple, siunitx}       % professional-quality tables
\usepackage{multirow}
\usepackage{amsfonts}       % blackboard math symbols
\usepackage{nicefrac}       % compact symbols for 1/2, etc.
\usepackage{microtype}      % microtypography
\usepackage{color}
\usepackage{caption}
\usepackage{adjustbox}
\usepackage{amsmath}
\usepackage{xspace}
\usepackage{soul}
\usepackage{subcaption}
\usepackage{graphicx}
\usepackage{hyperref}

\newcommand{\fax}{F$\alpha$X\xspace}

\def\DD{\mathcal{D}}
\def\FF{\mathcal{F}}

\title{Focus-and-Expand: Training Guidance Through Gradual Manipulation of Input Features}

\author{%
  Moab Arar \\
  Tel Aviv University \\
  \texttt{moabarar@mail.tau.ac.il} \\
  \And
  Noa Fish \\
  Tel Aviv University \\
  \texttt{noafish@gmail.com} \\
  \And
  Dani Daniel \\
  axilon \\
  \texttt{danid@axilion.com} \\
  \And
  Evgeny Tenetov \\
  axilion\\
  \texttt{evgenyt@axilion.com} \\
  \And
  Ariel Shamir \\
  The Interdisciplinary Center \\
  \texttt{arik@idc.ac.il} \\
  \And
  Amit Bermano \\
  Tel Aviv University \\
  \texttt{amberman@tauex.tau.ac.il}
}

\begin{document}

\maketitle

\begin{abstract}
 We present a simple and intuitive Focus-and-eXpand (\fax) method to guide the training process of a neural network towards a specific solution. Optimizing a neural network is a highly non-convex problem. Typically, the space of solutions is large, with numerous possible local minima, where reaching a specific minimum depends on many factors. In many cases, however, a solution which considers specific aspects, or features, of the input is desired. For example, in the presence of bias, a solution that disregards the biased feature is a more robust and accurate one. Drawing inspiration from Parameter Continuation methods, we propose steering the training process to consider specific features in the input more than others, through gradual shifts in the input domain. \fax extracts a subset of features from each input data-point, and exposes the learner to these features first, Focusing the solution on them. Then, by using a blending/mixing parameter $\alpha$ it gradually eXpands the learning process to include all features of the input. This process encourages the consideration of the desired features more than others. Though not restricted to this field, we quantitatively evaluate the effectiveness of our approach on various Computer Vision tasks, and achieve state-of-the-art bias removal, improvements to an established augmentation method, and two examples of improvements to image classification tasks. Through these few examples we demonstrate the impact this approach potentially carries for a wide variety of problems, which stand to gain from understanding the solution landscape.
\end{abstract}

% Introduction
\input{introduction}
% Related work
\input{previous}

 \bigskip 
% Method
\input{method}
 \bigskip 
% Experiments
\input{experiments}
 \bigskip 
% Discussion
\input{discussion}

 \bigskip 
 
\bibliographystyle{plain}
\bibliography{fax}

\appendix
\input{appendix_introduction}

\input{appendix_colored_mnist}

\input{appendix_data_augmentation}

\input{appendix_focusing}

\end{document}

%% file: introduction.tex
\section{Introduction}
\label{sec:introduction}

\input{figures/method_overview/overview}

When tackling a difficult search problem, such as locating our car in a parking lot, we usually start by searching for its color, especially if it is an uncommon one. If not, we focus on identifying its size or shape instead. The identification of the car may require examining several features, but we learn to leverage its distinctive features first to optimize the search. Even though the idea of \textit{feature search}~\cite{mcelree1999temporal} is intuitive in the context of human perception, it is typically ignored when considering the training of neural networks. As it turns out, a training process designed in this spirit can have several advantages.

Training a neural network involves minimizing a loss function that is, in general, a highly non-convex optimization problem. As a consequence, numerous local minima pose possible solutions, especially due to high number of parameters typical to deep neural networks. The specific solution reached during a training process depends on many factors, such as the representation and input-features used, the training data available, the order the samples are introduced to the learner, and more. Often the solution reached is considered robust because of the large size of the training data and randomness factors in the learning process. Still, there are many reasons why it would be desirable to direct the search towards a specific area in the optimization's solution landscape. Some of these reasons include efficiency or accuracy considerations, while others may involve avoiding biases or leveraging prior knowledge. 

In this paper, we present an intuitive and very simple to implement method to steer the solution towards a desired one, which we call \textit{Focus-and-eXpand} (\fax). We do this by proposing a training scheme that changes the fashion features are presented to the learner, and can be applied over any architecture. The key insight lies in the relationship between local minima (or solutions found by the training process) and features of the input. Given a training dataset $\DD$, in which each datapoint $d\in\DD$ is an instance of some input feature space $\FF$, we argue that in a myriad of cases a subset of features, $\FF_0 \subseteq \FF$, can be identified, that steers the solution towards a better one. Typically, $\FF_0$ would be overlooked during the training process, because other features, which are more easily found, distract the optimization. 

An example for such a case is when the training set of an image classification task includes large backgrounds that strongly correlate to the desired object each image is portraying. This setting drives the training process to a solution that examines the background to perform the classification, even though considering the object itself, albeit small, would clearly be more accurate. In such a situation, $\FF_0$ is the part of the image that depicts the actual object, and the background plays the role of the distracting element. When such conditions are identified, we propose exposing only $\FF_0$ to the learner at first (e.g. by cutting out the background). This enforces the optimization to \textbf{Focus} on a specific area in the solution landscape (e.g. solutions that examine the objects themselves). Then, we propose to gradually \textbf{eXpand} the learning process to include the rest of the features in $\FF$ (in our example -- the entire image). 

\fax draws inspiration from the general optimization approaches of parameter continuation. These methods deal with minimizing complex non-convex criteria~\cite{ContMethods-1980}, by defining a single-parameter family of criteria functions $L_\alpha(\theta)$. Continuation methods start with a simpler to optimize criterion $L_0(\theta)$, and gradually increase the parameter $\alpha$ while using the given solution $\theta$ of the previous iteration as the starting point for optimizing $L_\alpha(\theta)$, until reaching the desired criterion, $L_1(\theta)$. This formulation is used in the context of \fax as well. The optimization starts by optimizing using $\FF_0$ and then gradually increase the parameter $\alpha$ until reaching the full space $\FF$. In Section~\ref{sec:method}, we present two different methods to perform this continuation using a single parameter $\alpha$: by mixing the datasets from $\FF_0$ only with data from $\FF$, or by blending specific data instances.

Unlike the cases investigated by Feature Selection literature \cite{FS-challenge-04,FS-Survey-14,FS-Survey-16}, we consider the more common case for which using only $\FF_0$ is not the preferred choice (e.g. because the background can still help in distinguishing ambiguities, or because the segmentation of the object from the background is not available during inference time), hence the transition to $\FF$ is required. Moreover, we consider the typical case where a sharp transition, i.e. exposing $\FF$ (e.g.\ the entire image) immediately after training on $\FF_0$, would yield a poor outcome. This is common because the solution found by using $\FF_0$ alone would not handle the unseen features well, which would set the optimization free to restarts the learning process, without gaining from the local minimum the process was steered towards. Gradually eXpanding the learner to $\FF$ would, intuitively speaking, warrant small corrections to the solution which compensate for the introduced information, without straying too far from the initial local minimum found. 

\fax also seems to contradict the popular assumption that the right features should be chosen by the training process, and the more information exposed to it, the better. However, we note that the gradual introduction of information has already been proven effective in different scenarios. Examples include the hierarchical introduction of increasing image resolutions \cite{benaim2020structuralanalogy}, and Curriculum Learning approaches~\cite{CurrLearnBengio-09,weinshall2018curriculum,narvekar2020curriculum}. As we demonstrate in this paper, there are cases when judiciously exposing the learner to different features of the training data helps in solving bias problems, utilizing prior knowledge or improving training efficiency.

In Section~\ref{sec:experiments}, we demonstrate the flexibility of the proposed stepping through various computer vision tasks, even though the concept is not restricted to this regime alone. In already examined cases of a known bias in the training data \cite{BlindEye2018,LearnNot-19}, \fax is employed to encourage ignoring the correlating features from the final solution, improving upon the state-of-the-art. We also employ the \fax paradigm in the context of data augmentation, and demonstrate the improvement of an established strategy. Additionally, we demonstrate how \fax is used to Focus on specific features, i.e.\ to give them more importance in the decision making process. We demonstrate how \fax eliminates the need for test-time segmentation in a real-time classification task, by discouraging the use of image backgrounds. Finally, we also illustrate how Focusing the solution on the main object in a classification task helps improve the accuracy more than selecting which features the networks is exposed to during training.

Looking over the broad applicability of the concept, we are certain it can be employed in many other examples and ways, and hope this idea will inspire future works that develop \fax further.

%% file: figures/method_overview/overview.tex
\begin{figure}
    \centering
    \includegraphics[width=0.9\textwidth]{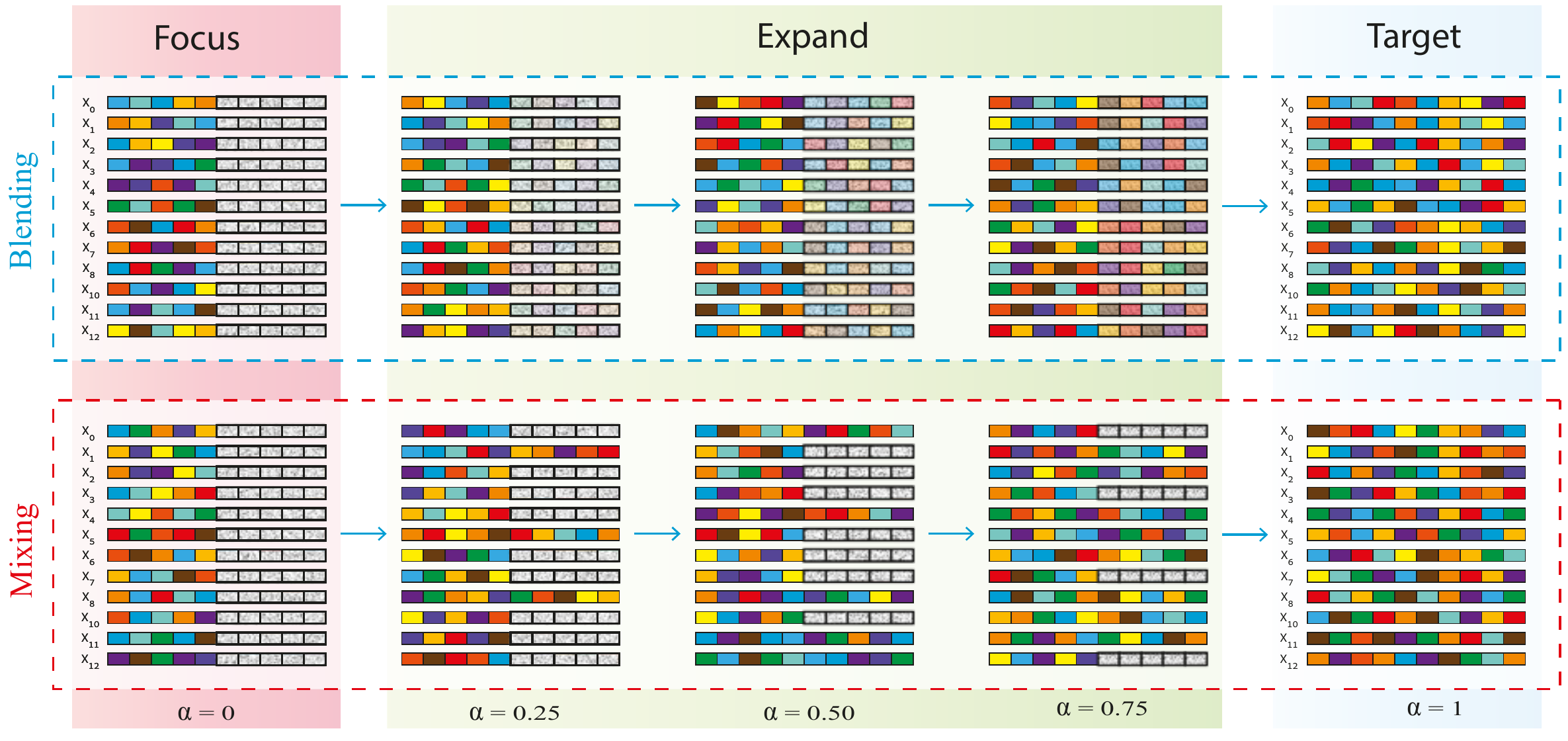}
    \caption{FaX overview. Initially, we focus on a subset of features in the dataset, only the first half of the features (less significant features can be, for example, noised out, zeroed, or completely removed). Then the data is gradually eXpanded by re-introducing all features back until the learner observes every feature in the data (last column). There are two methods for feature expansion, blending and mixing. In Blending implementation the features gradually fade-in via interpolation, while in Mixing an instance with all features present becomes more likely to be sampled during the training.}
    \label{fig:fax_illustration}
\end{figure}

%% file: previous.tex
\section{Related work}
\label{sec:previous}

Our work demonstrates the potential of input feature manipulation for robust learning of different tasks. In the following, we address the fields which we believe bear similarities to the \fax training framework.

\textbf{Bias removal} Even though the proposed concept could potentially be applied to many other areas, we focus on computer vision. A prominent example within the field, for which \fax can be used, is bias removal. The way \fax is employed in this context is described in Section~\ref{sec:experiments}. In known bias removal, the training set used by a network has been identified to include elements whose distributions are different than those during inference. to be able to generalize from the training bias to the test distribution, Alvi et al. \cite{BlindEye2018} propose a joint learning and unlearning approach whereby a classification model is trained such that its features representation disregards spurious information that exists in the data. 
Kim et al. \cite{LearnNot-19} define a regularization loss to minimize mutual information between feature embedding and target bias.
Teney et al. \cite{teney2020learning} pair minimally-different yet label-wise contrasting instances from the dataset, in order to steer the model away from spurious cues that may affect generalization.
Clark et al. \cite{clark2019don} train an ensemble of bias-only and robust models to encourage the robust model to gravitate toward data patterns that are potentially more general. All of these works, however, rely on additional networks, either of custom architecture that solves specific tasks, or ones that extract the biased feature from the data during training. We, on the other hand, propose only augmenting the input datapoints, hence do not employ any additional architecture, nor do we rely on derivable, or even automatic, means to extract the biased feature. Additionally, as we illustrate in Section~\ref{sec:experiments}, \fax outperforms the former two publications in an experiment suggested by Kim et al.~\cite{LearnNot-19}. Another popular approach used in the context of bias removal is domain adaptation.

\textbf{Domain adaptation} literature shows that the training process of various models can benefit from being initialized by training the given model on a similar task or dataset first. For example, Deng et at.~\cite{deng2018image} performs more robust re-identification by first learning to translate a given input image from the source domain to the target one, using one of the most popular image-to-image domain transfer tools -- CycleGAN~\cite{zhu2017unpaired}. Wang et al.~\cite{8836530} bridge the gap between a source and target domains by finding a common embedding for the two. In general, many of the works in this domain employ an adversarial component to bring the source and target distributions closer together \cite{tzeng2017adversarial,saito2018open,you2019universal}. For more in depth discussion, we turn the reader to surveys of the field \cite{wang2018deep,wilson2018survey}. Indeed our proposed framework can be cast to the domain adaptation world, as we also propose starting the training from a different dataset than the one ending the process with. However, none of these works offer a gradual, controlled, transition between the two domains, under the insight of what this transition offers the final model.

\textbf{Curriculum Learning} is a field that does propose the gradual introduction of information during training. Curriculum Learning (CL) techniques impose structure on the training data by sorting the samples from simple to hard, and using this ordering to progressively train the learner \cite{CurrLearnBengio-09}. 
This approach has been shown to accelerate convergence and improve performance of neural networks \cite{weinshall2018curriculum, PowerDT-19}.
For example, a method to automatically select the path of the curriculum to maximize learning has been proposed by Graves et al. \cite{graves2017automated}. Similarly, Zhang et al. \cite{zhang2019leveraging} leverage curriculum learning concepts to better training an objection detection network The benefits of curriculum learning have crossed the barrier of traditional tasks of computer vision, and have already been applied to tasks such as few-shots classification \cite{sun2019meta}, one-shot facial identification \cite{wu2019progressive}, video action detection \cite{yang2019step}, and in many other non-visual fields, such as Reinforcement Learning (RL) \cite{narvekar2020curriculum}. We draw inspiration from the literature of this field, and propose a principled guidance to the training process through controlling the information the network is exposed to. However, CL approaches control the information through ordered sampling of the input datapoints. In contrast, \fax extracts and controls the exposure of specific input features, without omitting datapoints in whole.

%% file: method.tex
\section{Method}
\label{sec:method}
Our method to steer solutions towards more desirable ones is a manipulation scheme on the input that applies simple operations to the features of input instances prior to their processing during training. 
%Denote the input feature space of our dataset as $\FF$. We formulate feature expansion (FE) by imposing structure on $\FF$.
%This structure, in its simplest form, defines a separation of $\FF$ into two sub-spaces, $\FF_0$ and $\FF_1$, such that $\FF_0 \cup \FF_1 = \FF$, but not necessarily $\FF_0 \cap \FF_1 = \emptyset$.
At the core of \fax is a gradual transition from learning based on $\FF_0$, which may be a simpler, or biased-reduced, feature sub-space, to learning based on $\FF$, which is the desired (e.g.\ more general or full) feature space.
This gradual shift is governed by the single parameter $\alpha$, which smoothly transitions from its initial value of 0 all the way to 1. When $\alpha=0$, the training process is exposed only to $\FF_0$, and when $\alpha=1$, it is exposed only to $\FF$. 
We provide two examples where this process can prove useful.
%thereby forming a family of input feature sub-spaces $\FF_\alpha$ that are sequentially shown to the model during training. 
%In the following we demonstrate several applications of this formulation.

\subsection{Use cases}

\textbf{Bias Removal.} Perhaps the most intuitive situation in which it is preferable to learn from one set of features instead of another is in the presence of bias. Training datasets are not always sufficiently balanced, general, and free from distractors that may impede the learning process. Specifically, often we find varying extents of bias within our training data, such that a set of features $\FF^-$ in the original feature space, $\FF$, is irrelevant to the task at hand, yet happens to be well correlated with the labeling. Examples include the appearance of cloudy skies when asked whether a person in the image is wearing a coat, or hair length in a gender identification task. In such cases, the learner is at risk of exploiting this bias and subsequently perform poorly at test time. 
To resolve these correlations, one option is to completely remove $\FF^-$ from the learning process. However, this not always an option, since it may not be possible during test time (e.g. cutting out the background in the former example would require first segmenting the image), or since $\FF^-$ may still improve performance (e.g. if the face is occluded in the latter example, the hair could be a strong hint). Hence, it is sometimes wiser to learn to handle these features with care, rather than to ignore them completely.
Using our formulation, we define $\FF_0 = \FF \setminus \FF^-$ and perform the \fax process. During the Focusing stage, \fax enforces the solution to consider only unbiased features. Then, the careful eXpansion from $\FF_0$ to $\FF$ enables the trained model to leverage the information available in $\FF^-$, while still giving $\FF_0$ higher priority. 
%Note that in order for this process to be successful, the performance offered by $\FF_0$ must be preferable to that offered by $\FF^-$, otherwise the optimization would just naturally go back to its old ways. In addition, note that if $\FF_0$ is already identified by the traditional training process, \fax would not hold merit.
%we opt to steer the learner away from resorting to a solution that exploits these irrelevant features, yet still be able to process them and show robustness when they appear. Using our formulation, we simply define $\FF_0 = \FF \setminus \FF^-$ and $\FF_1 = \FF$.

\textbf{Feature Focusing.} There are other reasons why one might choose to give some features more consideration than others, 
%Arik: I removed this because we don't show it at all...
%For example, one may choose to give race, color, or gender, less importance during a loan-granting decision making process. \fax could help in accommodating such considerations. By treating the undesired properties as bias, the previously described mechanism can be similarly employed. Another example 
for example, when involving prior knowledge. A learning task is often multi-layered and complex, forcing the optimization to search the solution landscape in many different locations and directions before finding the right ones during training. Thus, Focusing the network on the right features could save trial and error iterations, and significantly reduce training time, even if the final solution ends up to be the same with or without \fax. For instance, the color of the ink on a page is not relevant to OCR tasks, and neither does the shape of the room in which an action to be recognized is performed. 
%Be it to reduce wasted iterations, or even prejudice, 
\fax can be employed to Focus the learner on the relevant features in these cases. 

\subsection{Implementation}
To smoothly transition between $\FF_0$ and $\FF$, we implement a continuation through a family of criteria functions. 
%feature sub-spaces.  $\FF_\alpha$, we simplify our setting further. Instead of using $\FF_0$ and $\FF$, 
We separate (or augment) our training data $\DD$ into two groups $\DD_0$ and $\DD_1$, where $\DD_0$ are instances in $\DD$ that use only features in the sub-space $\FF_0$ (i.e.\ all features in $\FF \setminus \FF_0$ are given some neutral value), and $\DD_1$ are instances in $\DD$ that use all features in $\FF$. Based on these sub-sets we present two continuation strategies (see overview in Figure~\ref{fig:fax_illustration}):

\begin{itemize}
\item{\textbf{Mixing.}} 
This strategy exposes the two sub-groups of the training data $\DD_0$ and $\DD_1$ in a gradually increasing mixing ratio. While learning, we draw samples from each set with probability $1-g(\alpha)$ and $g(\alpha)$ respectively, for some monotonically increasing \textit{rate function} $g(\cdot)$. Intuitively speaking, the rationale for this strategy is that while the process is Focused on $\DD_0$, as $\alpha$ grows, it exposes elements from $\DD_1$ that would pull the optimization towards a final solution. The role of $\DD_0$ elements is to make sure the solution stays Focused, i.e. to enforce the solution back to the local minimum from which it came, while simultaneously evolving.
%$g(\alpha) = \alpha$  but we could define other monotonically increasing smooth functions from 0 to 1 using functions such as $tanh$ or $sigmoid$. \arik{Maybe need to formulate better?}

\item{\textbf{Blending.}} 
This strategy is only applicable when we have full bijective correspondence between the elements in $\DD_0$ to those in $\DD_1$. In this case, for each instance $x_0 \in \DD_0$ and its corresponding instance $x_1 \in \DD_1$, we define a new instance $x_\alpha \in D_\alpha$ as 
\[x_\alpha = (1 - g(\alpha))x_0 + g(\alpha)x_1.\]
In each step, we train the network using instances from $D_\alpha$. For the running example discussed in Section~\ref{sec:introduction}, this strategy means fading out the background completely for $\alpha = 0$, and gradually fading it back in as $\alpha$ grows. 

%\item{\textbf{Re-weighting.}} 
%This strategy is closest to the original parameter continuation optimization methods, as it uses the original loss function of the network, $L(\theta)$, to define a family of loss functions. Let $L_0(\theta) = L(\theta)|_{\DD_0}$ and $L_1(\theta) = L(\theta)|_{\DD_1}$. We now define \[L_\alpha(\theta) = (1 - g(\alpha))L_0(\theta) + g(\alpha)L_1(\theta).\]
%In other words, we have two loss criteria, where one is defined on the subset $\DD_0$ and the other on the subset $\DD_1$. During training, by using the parameter $\alpha$, we gradually move from measuring the loss on $\DD_0$ to measuring the loss on $\DD_1$, thus creating a family of loss functions.

\end{itemize}

\textbf{Continuation Rate.} 
In our experiments, we investigate the effects the choice of $g(\cdot)$ has on the learning process.
%In Section~\ref{sec:experiments}, we investigate the effects the choice of $g(\cdot)$ has on the learning process. 
%In Section~\ref{sec:experiments}, we investigate the effects of different choices of $g(\cdot)$. 
The simplest case is a \textbf{step} function, which means $\DD_0$ is initially given to the learner, until at some point we switch to using only $\DD_1$. This scenario is the one investigated by the domain adaptation literature. This is a well established approach with proven success, however we demonstrate that this strategy may yield inferior results. Intuitively, this is the case since when crossing the step in the function, the local minimum found during the first phase is not maintained, as discussed in Section~\ref{sec:introduction}. At the other is the \textbf{linear} function $g(\alpha)=\alpha$. While this indeed ensures a smooth transition, it turns out from our experiments that it may not be the best choice. Further details and analysis are provided in the appendix (\ref{sec:mnist_decaying_policies} and \ref{sec:data_aug_decaying_policy}). %that the mid-values of alpha are not challenging, and transitioning between the $[0.2,0.8]$ range could be done very quickly without compromising quality. This means that a linear approach is typically somewhat wasteful in terms of training time. Lastly, we have investigated using a \textbf{sigmoid} function, which transitions through the mid-values quickly, but gives more attention to the learning steps at both ends. Further details could be found in the supplementary. %In our experiments, this approach yielded the best results.

%\amit{There is some hidden text here discussing regularization. I think it's a good idea, but we cannot add it in without experiments}
% \textbf{Regularization.}  In practice, the $\alpha$ transation is a discrete, meaning that we change $\alpha$ at constant intervals (usually epochs). This in-turn will result in a non-smooth transition, which could result the network parameters to be shifted largely between epochs. Furthermore, increasing $\alpha$ could still lead to a trivial (unwanted) solution. In order to avoid that, we may add a regularization term where we restrict the network parameters from largely deviate between two consecutive epochs.  \moab{I think it might be necessary to add regularization term in some cases. This is still under investigation}

%% file: experiments.tex
\section{Applications}
\label{sec:experiments}
In this section, we analyze the ability of \fax to tackle different problems in computer vision, although we hope that the method will have broader implications. In particular, we show how \fax can be adapted to resolve biased data, where the training set suffers from severe (known) bias. Furthermore, we present examples where \fax is utilized to help focus the learner on certain features that ultimately promote better generalization at test time. For all discussed experiments, we offer more details in the appendix, including training statistics, visual examples from the datasets, and qualitative evaluations. 

In all experiments, unless otherwise stated, we use a Stochastic Gradient Descent (SGD) optimizer with momentum 0.9~\cite{SGDMOMENTUM}. We further set the weight decay regularization factor to $5e-4$.

%\subsection{Bias Removal}

\subsection{Colored MNIST}

\input{tables/mnist_accuracy}

\paragraph{Background} Naturally, models trained on biased datasets generalize poorly to instances drawn from the target data distribution. Nonetheless, in the following experiment we show that carefully employing our framework achieves a substantial increase in model generalization and accuracy. 

\paragraph{Dataset} We evaluate \fax on the colored MNIST dataset, as proposed by previous bias removal work~\cite{LearnNot-19}. The dataset is similar to MNIST~\cite{ORIGINAL_MINST}, with the exception that instances are colorized. The training set is color biased, i.e. samples belonging to the same class $C_{i} \in \{0,\dots{} 9 \}$ are synthesized with a color drawn from a normal distribution with mean $\mu_{i}$ and variance $\sigma^{2}$. The test set, on the other hand, is bias-free and random colors with mean $\mu = \sum_i \mu_i / 10$ and variance $\sigma^2$ are used to synthesized test instances. Different variance values are available in the dataset, ranging between $0.02$ and $0.05$, where the larger the variance is the less biased the training set is.
\textbf{Focus:} in order to remove bias from the training set, we convert all training data into grayscale images. This leads the network to focus on shape-related features such as edges, and ignore color, which is the biased feature. 
\textbf{eXpand:} since the real data is colored instances of MNIST~\cite{LearnNot-19, ORIGINAL_MINST}, using a network that is trained only on the source distribution (grayscale image) would generalize poorly. Therefore, gradually expanding our input-feature space to include color features is required. Note that one could employ a domain adaptation solution, where during test time input images undergo a translation from the target to the source distribution (i.e. be turn to gray) however our method eliminates this requirement, which could be strenuous in test-time. 
%\amit{there was something about being more accurate with color somehow, no?}
% \textbf{Experiment setup:} to test our method, a simple classification network~\cite{LearnNot-19} was trained for 100 epochs, and was optimized via a Stochastic Gradient Descent (SGD) with momentum 0.9. We use a (fixed) $0.01$ learning rate, with $5\exp{-4}$ factor for weight decay. Unless otherwise stated, the decay parameter $\lambda$ is gradually increased by $0.01$ at the end of each epoch (i.e, a linear increase from 0.0 to 1.0). We compare our method with two previous methods that were specifically devised to handle bias-removal~\cite{BlindEye-18, LearnNot-19}. We further train two baseline methods - one model is fed the original training data, and the other is trained on the grayscale converted images.

\paragraph{Experiment setup} to better evaluate our method, the same classification network~\cite{LearnNot-19} was used in all experiments. The training is performed for 100 epochs, with (fixed) $0.01$ learning rate, and the decay parameter $\alpha$ is gradually increased by $0.01$ at the end of each epoch from 0.0 to 1.0. We compare our method to the published results of two previous methods, who were specifically devised to handle bias-removal~\cite{BlindEye2018, LearnNot-19}. We further train two baseline methods - one model is fed the original training data, and the other is trained on the grayscale converted images. The training parameters for previous methods follow the setting reported in~\cite{LearnNot-19}. 

\paragraph{Results} Table~\ref{table:colored_mnist_accuracy} demonstrates better handling of bias by \fax compared to the baselines and previous works. Note also that contrarily to the previous methods, when using \fax there is no need to train specific networks for recognizing the bias, but only build appropriate training sets, which is arguably simpler. We also compare the effects of training the network with different values of $\alpha$, and advocate for the necessity to dynamically change $\alpha$ during the training (\ref{sec:mnist_decaying_policies}).

\subsection{Data augmentation - revisited}

\paragraph{Background} Data augmentation is a well established approach to training higher quality models that are better at generalizing to unseen examples. 
%In our formulation, one can view the original, un-augmented data as a form of biased data, and the augmented set as a version of the data where the bias is removed or reduced. This is because the data at test time does not necessarily display the augmentation properties, while at train time it is actively generated as such.
Often, however, the augmentation brings a form of non-realism into the mix, one that does not characterize test time instances. We enlist two successful data augmentation approaches for image classification, Mixup \cite{zhang2017mixup} and CutMix \cite{yun2019cutmix}, and follow their data preparation instructions of blending images. Despite the increase in performance that follows from applying these methods, we explore the effect of applying them in conjunction with \fax. We observe an increase in performance, demonstrating the merit in gradually steering the model back to the real, expected distribution. 

\paragraph{Dataset} This experiment is carried out on the CIFAR-10 and CIFAR-100 datasets~\cite{krizhevsky2009learning}. \textbf{Focus:} The focused data is the augmented set. This set is more general and diverse than the original set, and has been already proven to yield better results, thus we opt to begin the training with an exposure to this data. \textbf{eXpand:} The augmented data in this case is contrived, in that, at test time, images featuring elements that are mixed will not appear. Rather, test time images tend to resemble those in the original datasets. Thus, here, we gradually expand the exposure to the original data, letting the model become better tuned to the distribution expected at test time. The decay parameter $\alpha$ is the probability that an instance from the original dataset will be drawn, and, accordingly, $1-\alpha$ is the probability that an augmented instance is drawn. By setting $\alpha=0$, we begin by using augmented instances exclusively, similar to \cite{zhang2017mixup} and \cite{yun2019cutmix}, and as we gradually shift to $\alpha=1$, we increase the exposure to the original data.

\paragraph{Experiment setup} our experiment setup is similar to that in~\cite{zhang2017mixup}, but we have increased the number of epochs to 250 epochs. We examine two augmentation methods, Mixup~\cite{zhang2017mixup} and CutMix~\cite{yun2019cutmix}, that spawned four experiments: baseline with PreAct ResNet-18~\cite{he2016deep, PreActResnet}, baseline + \fax with PreAct ResNet-18, baseline with PyramidNet-200~\cite{han2017deep} and baseline + \fax with PyarmidNet-200 (where baseline follows the default training scheme used in either Mixup~\cite{zhang2017mixup} or CutMix~\cite{yun2019cutmix}). The initial learning rate was set to 0.2, and was reduced by a factor of 10 after 100 epochs, and again after another 50. The batch size was set to 128. The actual implementation of \fax in this experiment follows the Mixing implementation, where the decay parameter $\alpha$ was linearly increased from 0 to $k$ along the first 200 epochs, and then linearly increased from $k$ to 1 along the remaining 50, where $k=0.5$ for Mixup+\fax and $k = 0.25$ for CutMix+\fax. Further technical details about the implementation is provided in the appendix (see \ref{apendx:data_aug}).

\input{tables/cifar_accuracy.tex}
\paragraph{Results} Table~\ref{table:cifar_accuracy} displays the top-1 accuracy obtained on CIFAR10 and CIFAR100~\cite{krizhevsky2009learning}. As can be seen, by using \fax, better performance is obtained compared to the baseline methods. 
Note that the performance of Mixup~\cite{zhang2017mixup} is boosted when combined with \fax, and even achieves the highest accuracy on both datasets (among the tested methods).
% It is noteworthy that, from the Table~\ref{table:cifar_accuracy}, 
% the performance of Mixup~\cite{zhang2017mixup} was improved when combined with \fax, and it even gave the best accuracy on both datasets (among the tested methods).

%\subsection{Feature Focusing}

% Another application for which \fax is useful is termed feature focusing. Here we Focus the solution on specific features not because bias exists in the data, but rather because these features are favorable, e.g.\ because of prior knowledge, but are not easily found by the optimization process. 

\subsection{Classification of Abiotic Stress in Plants from RGB Images}

\input{tables/plants_accuracy.tex}

\paragraph{Background} This example deals with images of greenhouse plants treated in one of four fertilization policies. The task required is classification of the images into one of the four treatment policies. It turns out that a na\"ive training process tends to examine the entire image, even though the solution should consider the plant itself, rather than focus on distractors such as the background. Thus, we steer the solution toward specific features, i.e.\, the foreground elements, thereby assisting the optimization via feature focusing.

\paragraph{Dataset} the dataset is composed of 2874 training images and 1199 test images of 51 banana plants (for each class) of various ages (plant age varies between 0-61 days). The plants were segmented using a pre-trained network~\cite{SegmentationKimmel2019}, and a segmentation mask for the leaves is available for test and training images (see \ref{apendx:agrinet} for further details). \textbf{Focus:} we start by focusing the training on the segmented images, hence ignoring other parts of the image. The idea is that the background is irrelevant for the classification, and we expect the meaningful information to be within the plant itself. \textbf{eXpand:} we gradually expand the input features by reintroducing the background in the training set using either the Mixing or Blending technique.

\paragraph{Experiment setup} we use ResNet18~\cite{he2016deep} network for all our experiments, which was pre-trained on ImageNet~\cite{ImageNET}. The learning rate was fixed to $1e-3$, and the learning process lasted for 24 epochs, with batch size 32. 

\paragraph{Results} In Table~\ref{tabel:plants_test}, we report the test accuracy with respect to types of test images - segmented images (without background) and the original test images (with background). As expected, training with segmentation yields the best results, however, when introducing non-segmented images on a network that was not trained on images with a background, it fails to perform the classification. Training on the whole image on the other hand, yields inferior test accuracy when tested both on segmented and non-segmented data. Re-investigating Table~\ref{tabel:plants_test}, clearly shows that using \fax we are able to train our network to handle non-segmented images without compromising test accuracy. Full analysis is provided in the appendix, along with GradCAM~\cite{GradCAM} visualization, qualitatively demonstrating satisfactory results for the localization of the network when trained with \fax (see \ref{apendx:agrinet}).

%% file: tables/mnist_accuracy.tex
\begin{table}
    \centering
    \caption{Accuracy report on colored-MNIST\cite{LearnNot-19}. Colored (Gray) are baseline methods where the network trained on colored (grayscale version) images from the biased training test. The last two rows report the accuracy of our method, for each one of the implementations. We report the average and standard deviation of the method's accuracy, as observed by five different runs. The Gray baseline was tested on grayscale images, and the accuracy reports of the baseline methods is as reported by~\cite{LearnNot-19}.}
    \begin{adjustbox}{width=\textwidth}
    \begin{tabular}{lccccccc}
    \addlinespace
        \toprule [1.5pt]
        \multirow{2}{*}[-.5em]{Method} & \multicolumn{7}{c}{Training bias ($ \sigma^2) $} \\
        \cmidrule(lr){2-8}
        {} & .020 & .025 & .030  & .035 & .040 & .045 & .050 \\
        \addlinespace[2pt]
        \midrule[1.0pt]
        \addlinespace[4pt]
        Colored & .404 & .518 & .611 & .662 & .734 & .809 & .853 \\
        \addlinespace[4pt]
        Gray & .820 & .870 & .901 & .931 & .938 & .958 & .965 \\
        \addlinespace[4pt]
        BE~\cite{BlindEye2018} & .674 & .712 & .788 & .820 & .863 & .892 & .916\\
        \addlinespace[4pt]
        LNTL~\cite{LearnNot-19} & .818 & .8854 & .913 & .930 & .940 & .955 & .961 \\
        \addlinespace[2pt]
        \midrule
        %\addlinespace[4pt]
         %Loss cont. & $ .792 / .004 $ & $ .847 / .015$  & $.880 / .006 $ & $.916 / .003$ & $.927 / .001$ & $.947 / .002$ & $.959 / .002$ \\
         \addlinespace[4pt]
         Mixing & $ .852 / .011 $ & $ \textbf{.902 / .010}$  & $.912 / .007 $ & $\textbf{.941 / .002}$ & $.949 / .005$ & $\textbf{.962 / .002}$ & $.968 / .001$ \\
         \addlinespace[4pt]
         Blending & $ \textbf{.864 / .014} $ & $ .897 / .011$  & $\textbf{.919 / .007} $ & $.938 / .001 $ &  $\textbf{.952 / .000}$ & $.961 / .003$ & $\textbf{.970 / .002}$ \\
        \bottomrule [1.5pt]
        \addlinespace
    \end{tabular}
    \end{adjustbox}
\label{table:colored_mnist_accuracy}
\end{table}

%% file: tables/cifar_accuracy.tex
\begin{table}
    \centering
    \caption{Top-1 accuracy report on CIFAR10/CIFAR100~\cite{krizhevsky2009learning} datasets for different data augmentation techniques. For the PreAct ResNet18~\cite{PreActResnet} network, we report the average and standard deviation (avg./stdev) as observed in 5 different runs. Due it's lengthy training time, we report the best  accuracy of PyramidNet200~\cite{han2017deep} out of 2 different runs.}
    \label{tabel:cifar_accuracy}
    \begin{adjustbox}{width=\textwidth}
    \begin{tabular}{lccccccccc}
        \addlinespace
         \toprule[1.5pt]
          {} &\multicolumn{4}{c}{\textbf{CIFAR10}}&{}&\multicolumn{4}{c}{\textbf{CIFAR100}}\\
          \cmidrule(r){2-5}\cmidrule(r){7-10}
          {} &\multicolumn{2}{c}{Mixup~\cite{zhang2017mixup}} & \multicolumn{2}{c}{CutMix~\cite{yun2019cutmix}} & {} & \multicolumn{2}{c}{Mixup~\cite{zhang2017mixup}} & \multicolumn{2}{c}{CutMix~\cite{yun2019cutmix}}\\
          \cmidrule(r){2-3}\cmidrule(l){4-5}\cmidrule(r){7-8}\cmidrule(l){9-10}
          Network & baseline & \fax & baseline & \fax & {} & baseline & \fax & baseline & \fax \\
         \midrule[1.0pt]
         \addlinespace[4pt]
%--------------------------------------------- mixup baseline --- mixup ours  --- cutmix baseline --- cutmix ours  ----- mixup baseline --- mixup ours  --- cutmix baseline --- cutmix ours  --- 
         PreAct-ResNet18~\cite{PreActResnet} & $95.96 / .06$ & $96.1 / .1$ & $96.27 / .1$  & $96.29 / .1$  & & $78.26 / 0.35$ & $79.60 / 0.15$ & $79.78 / 0.2$ & $80.22 / 0.21 $\\
         \addlinespace[4pt]
         PyramidNet200 & 97.36 & \textbf{97.47} & 97.4  & 97.42 & & 83.13& \textbf{84.51} & 84.14  & 84.39 \\
         \bottomrule[1.5pt]
         \addlinespace
      \end{tabular}
      \end{adjustbox}
\label{table:cifar_accuracy}
\end{table}

%% file: tables/plants_accuracy.tex
\begin{table}
    \centering
    \caption{Top-1 accuracy report on fertilization stress test. We report the best out of three accuracy results obtained for each one of the three training policies for ResNet18~\cite{he2016deep}. The network is either trained with the original data (without background - With BG), the data after background removal (W/O BG) and finally using our method \fax (in both implementations). The rows represents the accuracy of the network on the original data with the background (With BG), and instances where the background was discarded (W/O BG).}
    \label{tabel:plants_test}
    \begin{adjustbox}{width=0.8\textwidth}
    \begin{tabular}{lcccc}
        \addlinespace[1.25pt]
         \toprule[1.5pt]
          {} &\multicolumn{4}{c}{\textbf{Training Policy}}\\
          \cmidrule(r){2-5}
          Test set & With BG & W/O BG & \fax - Blending & \fax - Mixing\\
         \midrule[1.0pt]
         \addlinespace[4pt]
         With BG & 59.29 & 33.02 & 62.22  &  64.55\\
         \addlinespace[4pt]
         W/O BG  & 58.79  & \textbf{65.42} & 48.87  & 54.12 \\
         \bottomrule[1.5pt]
         \addlinespace
      \end{tabular}
      \end{adjustbox}
\end{table}

%% file: discussion.tex
\section{Discussion}
\label{sec:discussion}

We have presented \fax, a method to improve the generalization of deep neural networks that steers solutions towards more desirable ones by manipulating the input features.
Our main insight is that we can steer the training solution to a local minima, and encourage it to stay in its proximity by careful and gentle gradual shifts.
We demonstrate the wide applicability and simplicity of the concept: 
%and shows how a thorough understanding of a given problem, and especially the behavior of the data and the employed network, can be leveraged to easily and intuitively improve performance.
we improve the performance of various established architectures and methods using only approximately two dozen lines of code. 

Importantly, our method and evaluation validate the claim that the solution found at the end of a training process depends on the order of data (and features) it is exposed to, and thus can be controlled. We do note, however, that finding the right way to employ \fax requires careful tuning. 
The key lies with understanding what the network learns from the focusing dataset $\DD_0$. For example, removing the background of images in the abiotic stress classification task described in Section~\ref{sec:experiments}, teaches the network to focus on the leaves to perform the classification. On the other hand, removing the background in an object segmentation task would probably not prove helpful at all. Additionally, one must keep in mind that a successful focusing process must end up in a local minimum. Hence, focusing the network on significantly inferior aspects of the input (e.g.\ on just the nose in a facial identification task) would prove futile, since eXpanding would would immediately cause the optimization to choose more beneficial features. Similarly, focusing the network on features which are already active in the na\"ively found solution is just as fruitless.

Our proposed approach leaves much room for future investigation. For one, we would  like to continue and identify more examples and tasks, possibly outside Vision, which can benefit from our intuitive and simple modification. These examples could potentially be applied to any machine learning scheme, including language processing, graph embeddings, or even non-neural learning methods. Extending the idea further, it is worth exploring whether \fax could be iteratively employed, to form a scheme more similar to the multi-step exposition of information approaches seen in curriculum learning.  

\textbf{Broader Impact.} As we have demonstrated, our method could be applied for bias removal. We believe this statement holds also for handling many forms of bias, and not only those that hamper performance. Such cases could be designed to include social or ethnicity based biases introduced during the learning process. Furthermore, Controlling the features a network is using for a given solution could potentially be useful also for enhancing explainability - where the network is pushed towards using more human legible considerations.

In conclusion, we believe the proposed method offers means to gain meaningful insights regarding the manner in which networks train and behave, and hope to see it extensively further developed.

%% file: appendix_introduction.tex
\section{Experiments summary}
The experiments' setup are summarized in Table~\ref{table:expirments_summary}, and further details are provided in later sections. Furthermore, an in-depth analysis is provided for each experiment, including the effect of using different decaying policies on the overall performance of the model.

\input{tables/expirement_summary/table}

%% file: tables/expirement_summary/table.tex
\begin{table}[ht]
    \centering
    \caption{Summary of experiments. Full details can be found within the document.}\label{table:expirments_summary}
        \begin{filecontents*}{tables/expirement_summary/exp_summary.csv}
Experiment name,Optimizer,Opt. Param.,lr,lr-decay,Batch size,Weight-decay,Epochs,Network,Preprocess
Colored-MNIST-1,SGD,Momentum 0.9,0.1,None,128,0.0005,100,Simple,Pre-1
Colored-MNIST-2,SGD,Momentum 0.9,0.01,None,128,0.0005,100,Simple,Pre-1
Augmenetation-1,SGD,Momentum 0.9,0.1,MultiStepLR,128,0.0005,250,PreAct-ResNET-18,Pre-2
Augmenetation-2,SGD,Momentum 0.9,0.1,MultiStepLR,128,0.0005,250,Pyramid-200,Pre-2
ABIOTIC-STRESS,SGD,Momentum 0.9,0.001,None,32,0.0005,24,Pretrained ResNET18,Pre-1
Hands-Classification,Adam,betas: 0.9/0.999,0.0001,None,32,0.000001,35,ResNET-18,Pre-1
\end{filecontents*}
\csvreader[
            tabular={lcccccccc},
            table head=\toprule\bfseries Experiment name & \bfseries Optimizer & \bfseries Optim. params. & \bfseries lr & \bfseries lr-decay & \bfseries Batch size & \bfseries Weight decay & \bfseries Epochs & Network \\\midrule,
            table foot=\bottomrule,
             before reading=\begin{adjustbox}{width=\textwidth},
            after reading=\end{adjustbox}]
            {tables/expirement_summary/exp_summary.csv}
            {1=\name,2=\optim,3=\optimparam,4=\lr,5=\lrdecay,6=\bsize,7=\weightdecay,8=\epochs,9=\network}
            {\name & \optim & \optimparam & \lr & \lrdecay & \bsize &  \weightdecay & \epochs & \network }
       % \end{adjustbox}
\end{table}

% Weight-decay,Epochs,Network,Preprocess

%% file: appendix_colored_mnist.tex
\section{Colored MNIST}
\label{sec:mnist}
\input{tables/mnist_full_report/blending}

In this section, we provide further analysis on the bias removal experiment. The experiment setup and parameters are summarized in Table~\ref{table:expirments_summary}. For \fax, we use the setup denote by Colored-MNIST-1, and for other experiments we run with setup denoted by Colored-MNIST-2. The network we used (denoted as Simple in the table), is the implementation used by Kim et al.~\cite{LearnNot-19} and is available in the authors github repository\footnote{\url{https://github.com/feidfoe/learning-not-to-learn}}. 

Furthermore, in Table~\ref{table:blending_mnist}, we report the accuracy obtained by a set of experiments, from which we derived the avg. and std. top-1 accuracy reported in the paper. 

Finally, in Figure~\ref{fig:colored_mnist}, we present sample instances from the training and test datasets with color variance $\sigma^2 = 0.02$, which indicates highly biased sets. The figure clearly demonstrates the color bias in the training set.

\input{figures/mnist_dataset/mnist_dataset_visualization}

\subsection{Decaying Policies}
\label{sec:mnist_decaying_policies}

We investigate the effect of employing different decay policies for the $\alpha$ parameter. Specifically, we focus on the following three policies: 

\begin{enumerate}
    \item \textbf{Linear}: the decay parameter $\alpha$ is increased by $1/\texttt{epochs}$ every epoch, where $\texttt{epochs}$ is the total number of training epochs.
    \item \textbf{Constant}: the decay is fixed and set to a constant $\alpha$ during training. The constant values used are $\alpha \in \{0.25, 0.50, 0.75, 1.00\}$.
    \item \textbf{Step}: this is equivalent to domain adaption methods, where the decay parameter $\alpha$ is set to zero up to a predetermined $\texttt{epoch}$, from where it is increased to $1.0$.
\end{enumerate}
 
In Figure~\ref{fig:mnist_compare_with_constant}, we show the test accuracy as a function of the current epoch for linear and constant decaying policies. We report the accuracy for the two versions -- Mixing and Blending. Note that for models trained on grayscale images only, we report the test accuracy for a grayscale version of the test set (as proposed by Kim et al.~\cite{LearnNot-19}).

\input{figures/mnist_constant_alphas/colored_mnist_constant_alphas}

We observe that the Mixing strategy yields consistent results across the board, whereas the Blending strategy clearly struggles under large values of $\alpha$. We postulate the reason stems from the fact that Blending generates instances that are a hybrid of the original (biased) image, and its biased-removed version. This means that at every step, the network is required to adapt to a different data distribution, which might be disadvantageous in this case. In contrast, the Mixing strategy simply selects one version or the other, and does not attempt to create a mixture, thus creating a more consistent set of instances at every step of the way. This suggests adapting $\alpha$ during training yields best accuracy, and is crucial for training with the Blending strategy (this is because the network could derive the bias from input, since the blending is kept constant, and the network could filter out the unbiased part - just as in noise canceling networks).

In Table~\ref{table:mnist_compare_with_step}, we report the accuracy obtained when training on different step functions, or in other words, we ask how a domain adaptation approach would perform using our suggested approach. Our two strategies clearly yield better performance, proving confirmation to our claim that small changes to $\alpha$ encourage the optimization to remain in the proximity of the local minimum found, while a large change lets the learning process stray further away from the solution found while using $\alpha=0$. Note also the gap between the three step functions, where applying the step at epoch 25 is more favorable to applying it later on at epoch 50 or 75. This supports our claim even further, since a lighter initial training process (of 25 epochs) means the weights are not as tuned already to the local minima, and hence leaving it when exposed to $\alpha=1$ is easier.
%A possible reason for that may be the small remaining bias that still exists in the bias-removed data (see Figure \ref{fig:colored_mnist}, where one can note the differences in luminance between the grayscale digits, resulting from differences in luminance in the original RGB colorspace). The existence of this small bias in combination with a longer exposure (epoch 50) to it, before reverting back to the fully biased dataset, possibly leads to a more pronounced overfit to the bias-removed instances, with subsequent difficulties to then handle the varied data at test time.

% \input{figures/mnist_compare_with_step/compare_with_step}
\input{tables/mnist_table/step_table}

%% file: tables/mnist_full_report/blending.tex
\begin{table}
    \centering
    \caption{\fax results on Colored MNIST for different runs. On the left (resp. right), we report the accuracy obtained for Blending (resp. Mixing) implementation. The columns represent an experiment with specific training bias ($\sigma^2$).}\label{table:blending_mnist}
        \begin{filecontents*}{tables/mnist_full_report/blending_report.csv}
,a,b,c,d,e,f,g
1,0.8785,0.903,0.9134,0.9402,0.9524,0.9624,0.9686
2,0.8687,0.911,0.9237,0.9376,0.9513,0.9561,0.971
3,0.8533,0.9004,0.9155,0.9368,0.9518,0.9641,0.9691
4,0.8455,0.8951,0.9132,0.9376,0.9523,0.9614,0.9695
5,0.8743,0.8796,0.9312,0.9405,0.9529,0.9634,0.9742
\end{filecontents*}
\csvreader[
            tabular={ccccccc},
            table head=\toprule0.02 & 0.025 & 0.03 & 0.035 & 0.04 & 0.045 & 0.05 \\\midrule,
            table foot=\bottomrule,
            before reading=\begin{adjustbox}{width=0.44\textwidth},
            after reading=\end{adjustbox}]
            {tables/mnist_full_report/blending_report.csv}
            {1=\aa,2=\bb,3=\cc,4=\dd,5=\ee, 6=\ff, 7=\gg, 8=\hh}
            {\bb & \cc & \dd & \ee &\ff & \gg & \hh}
            \quad
            \begin{filecontents*}{tables/mnist_full_report/mixing_report.csv}
x,a,b,c,d,e,f,g
1,0.8714,0.9065,0.9146,0.9437,0.9507,0.966,0.9708
2,0.8406,0.9108,0.9011,0.9434,0.9492,0.9591,0.9686
3,0.8535,0.8999,0.9088,0.9386,0.9406,0.9633,0.9703
4,0.8443,0.8858,0.9183,0.9441,0.9519,0.9629,0.9676
5,0.853,0.9083,0.9198,0.9393,0.9566,0.9612,0.9673
\end{filecontents*}
\csvreader[
            tabular={ccccccc},
            table head=\toprule 0.02 & 0.025 & 0.03 & 0.035 & 0.04 & 0.045 & 0.05 \\\midrule,
            table foot=\bottomrule,
            before reading=\begin{adjustbox}{width=0.44\textwidth},
            after reading=\end{adjustbox}]
            {tables/mnist_full_report/mixing_report.csv}
            {1=\aa,2=\bb,3=\cc,4=\dd,5=\ee, 6=\ff, 7=\gg, 8=\hh}
            {\bb & \cc & \dd & \ee &\ff & \gg & \hh}
\end{table}

%% file: figures/mnist_dataset/mnist_dataset_visualization.tex
\begin{figure}
    \centering
    \includegraphics[width=0.225\textwidth]{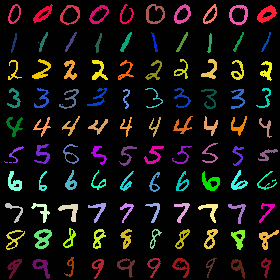}
    \includegraphics[width=0.225\textwidth]{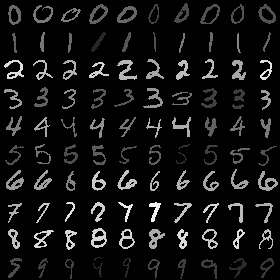}
    \includegraphics[width=0.225\textwidth]{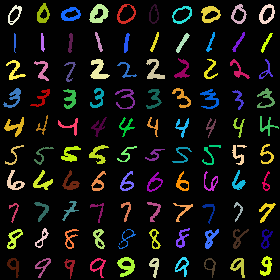}
    \includegraphics[width=0.225\textwidth]{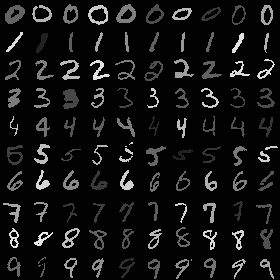}
    
    \caption{Colored MNIST dataset visualization for $\sigma^2 = 0.02$. Left-to-right: biased training set, training set in grayscale, test set and test set in grayscale.}
    \label{fig:colored_mnist}
\end{figure}

%% file: figures/mnist_constant_alphas/colored_mnist_constant_alphas.tex
\begin{figure}
    \centering
    \includegraphics[width=\textwidth]{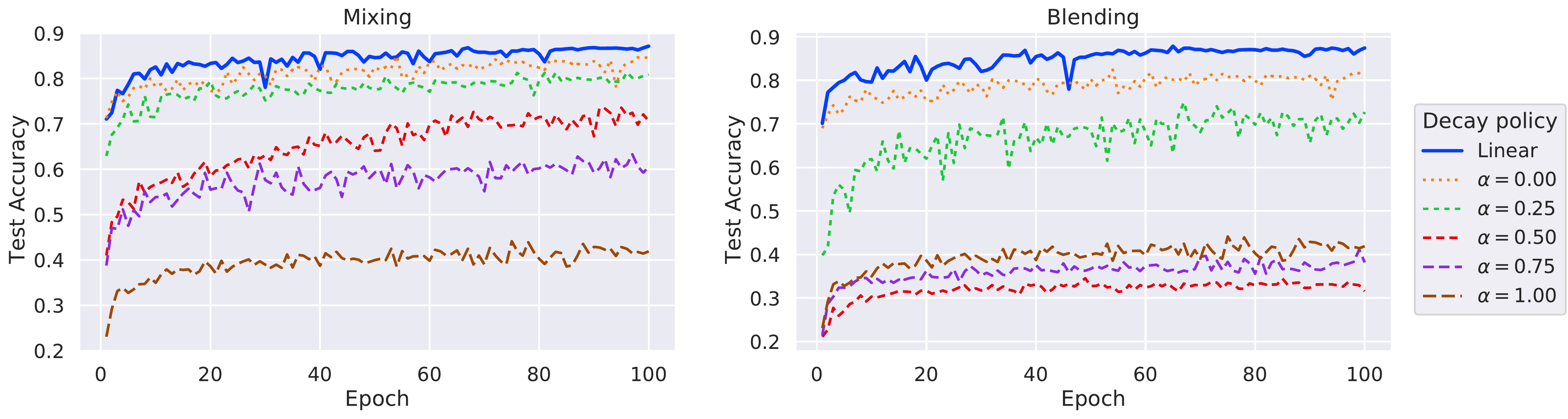}
    \caption{Test accuracies obtained by training with various constant $\alpha$ values, compared to our linear decaying policy. We show the two strategies -- Mixing (left) and Blending (right).}
    \label{fig:mnist_compare_with_constant}
\end{figure}

%% file: tables/mnist_table/step_table.tex
\begin{table}
    \centering
    \caption{Our linear decaying policy vs. the step (domain adaptation) policy. In the step policy, $\alpha$ is changed from $0.0$ to $1.0$ at either epoch $25$, $50$ or $75$. We report the best accuracy obtained during training, and the average accuracy obtained in the last 10 epochs.} \label{table:mnist_compare_with_step}
            \begin{filecontents*}{tables/mnist_table/mnist_step.csv}
-,Step 25,Step at 50,Step at 75,Mixing,Blending
Best,0.785,0.725,0.647,0.8714,0.8785
Avg.,0.767,0.702,0.628,0.866,0.87
\end{filecontents*}
\csvreader[
            tabular={lccccc},
            table head=\toprule {} & Step at 25 & Step at 50 &  Step at 75& \fax-Mixing & \fax-Blending\\\midrule,
            table foot=\bottomrule]
            {tables/mnist_table/mnist_step.csv}
            {1=\aa,2=\bb,3=\cc,4=\dd,5=\ee, 6=\ff}
            {\aa &\bb & \cc & \dd & \ee &\ff}
\end{table}

%% file: appendix_data_augmentation.tex
\section{Data augmentation}
\label{apendx:data_aug}
\input{tables/data_augmentation_cifar/cifar10}
\input{tables/data_augmentation_cifar/cifar100}

In this section, we provide further analysis on the data augmentation experiment. Similar to Section~\ref{sec:mnist}, a summary of the experiments setup are provided in Table~\ref{table:expirments_summary}. 

\paragraph{Implementation } our implementation is based on the official source code of the Mixup augmentation method~\cite{zhang2017mixup}\footnote{\url{https://github.com/facebookresearch/mixup-cifar10}}. We also integrate CutMix~\cite{yun2019cutmix} in the code using the official implementation of the CutMix method\footnote{\url{https://github.com/clovaai/CutMix-PyTorch}}. To support \fax, we adapt the Mixing strategy with minor changes: since both CutMix~\cite{yun2019cutmix} and Mixup~\cite{zhang2017mixup} perform the augmentation on each drawn batch, thus we also perform the Mixing strategy in batch-resolution. Meaning, we either chose an augmented batch or non-augmented batch. Note that since the dataset size is sufficiently large compared to the batch-size, and data is shuffled during training, we get that each instance is roughly drawn from $\DD_0$ with probability $1 - g(\alpha)$. Furthermore, in these methods, a mixing parameter $\lambda$ is used to control the amount the augmentation performed on the input image (e.g. for Mixup~\cite{zhang2017mixup}, $\lambda$ controls the interpolation performed on the data, where $\lambda=1$ means no augmentation). The parameter $\lambda$ is uniformly drawn from $(0,1)$, where in our implementation we use the variable $\lambda' = (1-\alpha)\cdot \lambda + \alpha$.

\subsection{Decaying policies}
\label{sec:data_aug_decaying_policy}
\input{figures/piecewise_linear/piecewise_linear}
As in section~\ref{sec:mnist_decaying_policies}, we compare different decaying methods with respect to the augmentation experiment as well. In all experiments for this section, we use \fax-Mixing with mixup~\cite{zhang2017mixup} augmentation as the baseline method, and the dataset used is CIFAR100. We denote $\texttt{max-epochs}$ as the total number of epochs, and $\texttt{e}$ as the current epoch. The policies we have tried are as follows:
\begin{enumerate}
    \item \textbf{Piecewise linear:} in this policy, $\alpha$ is linearly increased from 0.0 to $k_2$ in the first $k_1\cdot \texttt{max-epochs}$ epochs, and is then linearly increased to $1.0$ in the remaining $(1-k_1)\cdot \texttt{max-epochs}$ epochs. Mathematically, the function is given by:
         \[
             f_{k_1,k_2}(\texttt{e}) =
                 \begin{cases}
                          \frac{k_2}{k_1}\cdot \frac{\texttt{e}}{\cdot \texttt{max-epochs}} & \text{if $e\leq k_1 \cdot \texttt{max-epochs}$ }\\
                          k_2+\frac{1-k_2}{1-k_1}\cdot  (\frac{\texttt{e}}{\texttt{max-epochs}}-k_1) & \text{otherwise}\\
                 \end{cases}
          \]
    See Figure \ref{fig:piecewise_linear} for an example. Note that for $k_1 = k_2$, the decay is simply linear.
    \item \textbf{Step function:} in this policy, alpha is set to $0.0$ in  the first  $k_1\cdot \texttt{max-epochs}$ epochs, and to $k_2$ in the remaining $(1-k_1)\cdot \texttt{max-epochs}$ epochs.
\end{enumerate}

In our experiments, we consider different values of $k_1$ and $k_2$ for each of the two decay policies. We found it intuitive to classify each experiment into the following categories, based on the nature of the values of the parameters and the decay effect that they create:

\begin{itemize}
    \item \textbf{Fast start:} the value of $\alpha$ is rapidly increased in the first half of training.
    \item \textbf{Moderate start:} the value of $\alpha$ increases almost linearly. 
    \item \textbf{Slow start:} the value of $\alpha$ is conservatively increased, only to be rapidly increased to reach $1.0$ during the last epochs.
    \item \textbf{Step: } different step functions are examined, in particular, we set $k_2=1$ and consider different values of  $k_1 \in \{0.25, 0.5, 0.75\}$
    \item \textbf{Constant: } throughout training, the same composition of both the augmented and the original data is used. In particular, for these experiments, we employed a step function with $k_1=0$ and $k_2 \in \{0.25, 0.5, 0.75\}$
\end{itemize}

\input{figures/data_augmentation_summary/data_augmentation_summary}

In Figure~\ref{fig:data_augmentation_box_plot}, a box-plot chart summarizes our findings per category. Table~\ref{table:data_augmentation_alpha_policy} contains a detailed report of all the experiments we ran with their corresponding parameter values and classification into categories. Thus, Figure~\ref{fig:data_augmentation_box_plot} is essentially a visual summary of Table ~\ref{table:data_augmentation_alpha_policy}.
Examining the box-plot, we observe the superiority of the "Slow Start" category, which is our chosen approach. At the end of the spectrum, we find "Fast Start" and "Constant", the first of which exposes the original data too quickly resulting in loss of effect of the augmented data. The second ("Constant") makes no use of the idea of decay, demonstrating its usefulness by way of omission. 

\input{tables/data_augmentation_alpha_policy/table}

%% file: tables/data_augmentation_cifar/cifar10.tex
\begin{table}
    \centering
    \caption{\fax results on CIFAR10 over several runs. We report on each column the accuracy obtained for a specific training setup. The naming we use follow the convention [Augmentation Method]-[Network Number]-[\fax or baseline]. The augmentation considered are Mixup~\cite{zhang2017mixup} or CutMix~\cite{yun2019cutmix}. The networks are either N1 (for PreAct-Resnet18~\cite{PreActResnet}) or N2 for (PyramidNet200~\cite{han2017deep}).} \label{table:cifar10_resuts}
            \begin{filecontents*}{tables/data_augmentation_cifar/cifar10_results.csv}
ResNet18 - Mixup Ours,ResNet18 - Mixup Baseline,ResNet18 - Cutmix Ours,ResNet18 - Cutmix Baseline,PyramidNet Mixup Ours,PyramidNet Mixup Baseline,PyramidNet Cutmix Ours,PyramidNet Cutmix Baseline
95.99,96.01,96.29,96.33,97.47,97.36,97.42,97.4
96.07,95.89,96.33,96.38,97.43,97.18,97.41,97.3
96.29,95.98,96.18,96.12,-,-,-,-
96.01,95.91,96.2,96.3,-,-,-,-
96.14,96.03,96.45,96.23,-,-,-,-
\end{filecontents*}
\csvreader[
            tabular={cccccccc},
            table head=\toprule Mixup-N1-\fax & Mixup-N1 & CutMix-N1-\fax & CutMix-N1 & Mixup-N2-\fax & Mixup-N2 & CutMix-N2-\fax & CutMix-N2 \\\midrule,
            table foot=\bottomrule,
            before reading=\begin{adjustbox}{width=\textwidth},
            after reading=\end{adjustbox}]
            {tables/data_augmentation_cifar/cifar10_results.csv}
            {1=\aa,2=\bb,3=\cc,4=\dd,5=\ee, 6=\ff, 7=\gg, 8=\hh}
            {\aa &\bb & \cc & \dd & \ee &\ff & \gg & \hh}
\end{table}

%% file: tables/data_augmentation_cifar/cifar100.tex
\begin{table}
    \centering
    \caption{\fax results on CIFAR100. Similar naming convention was used as in Table~\ref{table:cifar10_resuts}.} \label{table:cifar100_resuts}
            \begin{filecontents*}{tables/data_augmentation_cifar/cifar100_results.csv}
ResNet18 - Mixup Ours,ResNet18 - Mixup Baseline,ResNet18 - Cutmix Ours,ResNet18 - Cutmix Baseline,PyramidNet Mixup Ours,PyramidNet Mixup Baseline,PyramidNet Cutmix Ours,PyramidNet Cutmix Baseline
79.84,78.66,80.42,79.52,84.16,83.11,84.24,83.99
79.57,78.46,80.31,79.9,84.51,83.13,84.39,84.14
79.62,77.99,80.38,79.6,-,-,-,-
79.44,77.95,79.89,79.97,-,-,-,-
79.55,78.61,80.22,79.9,-,-,-,-
\end{filecontents*}
\csvreader[
            tabular={cccccccc},
            table head=\toprule\bfseries Mixup-N1-\fax & Mixup-N1 & CutMix-N1-\fax & CutMix-N1 & Mixup-N2-\fax & Mixup-N2 & CutMix-N2-\fax & CutMix-N2 \\\midrule,
            table foot=\bottomrule,
            before reading=\begin{adjustbox}{width=\textwidth},
            after reading=\end{adjustbox}]
            {tables/data_augmentation_cifar/cifar100_results.csv}
            {1=\aa,2=\bb,3=\cc,4=\dd,5=\ee, 6=\ff, 7=\gg, 8=\hh}
            {\aa &\bb & \cc & \dd & \ee &\ff & \gg & \hh}
\end{table}

%% file: figures/piecewise_linear/piecewise_linear.tex
\begin{figure}
    \centering
    \includegraphics[width=0.5\textwidth]{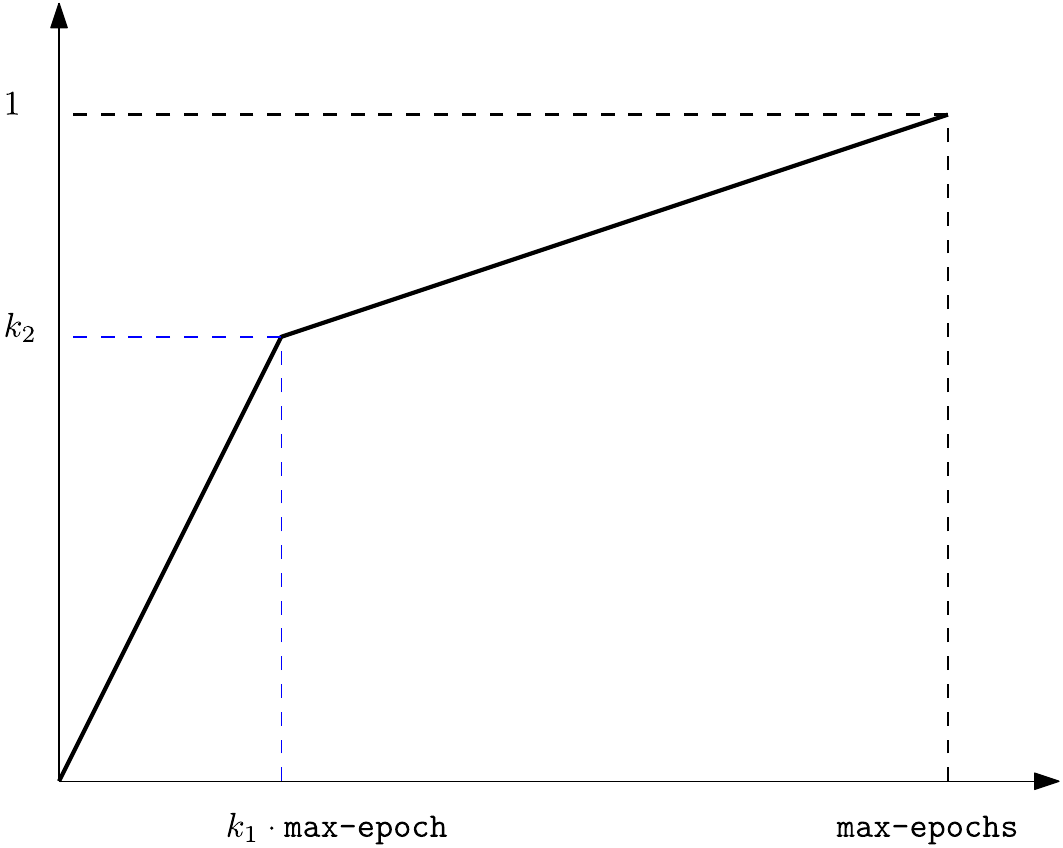}
    \caption{Piece-wise linear decay policy.}
    \label{fig:piecewise_linear}
\end{figure}

%% file: figures/data_augmentation_summary/data_augmentation_summary.tex
\begin{figure}
    \centering
    \includegraphics[width=\textwidth]{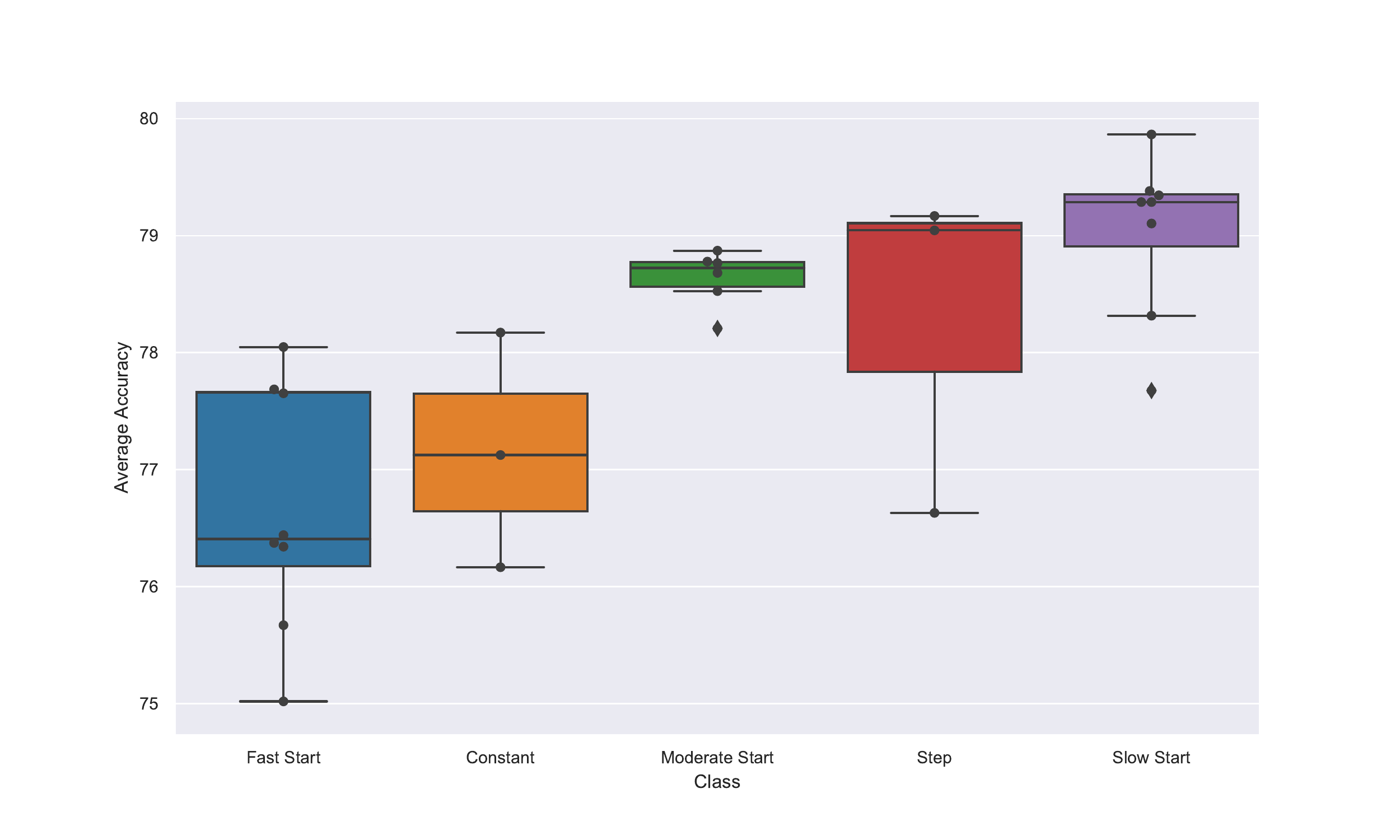}
    \caption{Comparison of different decay policies on the CIFAR100 dataset. The results are classified into five categories of policies, based on parameter values. The reported accuracy is the average top-1 test accuracy as observed over the final 10 epochs.}
    \label{fig:data_augmentation_box_plot}
\end{figure}

%% file: tables/data_augmentation_alpha_policy/table.tex
\begin{table}
    \centering
    \caption{Comparison of different decay policies on the CIFAR100 dataset.}\label{table:data_augmentation_alpha_policy}
        \begin{filecontents*}{tables/data_augmentation_alpha_policy/augmentation_data.csv}
,K1,K2,Type,Class,Average Accuracy,Best Accuracy
0,0.0,0.5,Piecewise linear,Fast Start,76.374,76.51
1,0.0,0.25,Piecewise linear,Moderate Start,78.207,78.4
2,0.0,0.75,Piecewise linear,Fast Start,75.671,75.77
3,0.0,1.0,Piecewise linear,Fast Start,75.020,75.23
4,0.5,0.0,Piecewise linear,Slow Start,79.345,79.53
5,0.5,0.5,Piecewise linear,Moderate Start,78.766,78.96
6,0.5,0.25,Piecewise linear,Slow Start,79.382,79.52
7,0.5,0.75,Piecewise linear,Fast Start,78.048,78.22
8,0.5,1.0,Piecewise linear,Fast Start,77.686,77.81
9,0.25,0.0,Piecewise linear,Moderate Start,78.872,79.04
10,0.25,0.5,Piecewise linear,Fast Start,77.653,77.79
11,0.25,0.25,Piecewise linear,Moderate Start,78.526,78.68
12,0.25,0.75,Piecewise linear,Fast Start,76.441,76.62
13,0.25,1.0,Piecewise linear,Fast Start,76.342,76.58
14,0.75,0.0,Piecewise linear,Slow Start,79.865,79.94
15,0.75,0.5,Piecewise linear,Slow Start,79.287,79.41
16,0.75,0.25,Piecewise linear,Slow Start,79.291,79.4
17,0.75,1.0,Piecewise linear,Moderate Start,78.778,78.94
18,1.0,0.0,Piecewise linear,Slow Start,77.676,78.5
19,1.0,0.5,Piecewise linear,Slow Start,79.104,79.27
20,1.0,0.25,Piecewise linear,Slow Start,78.316,79.29
21,1.0,0.75,Piecewise linear,Moderate Start,78.682,78.87
22,0.0,0.5,Step Function,Constant,77.125,77.55
23,0.0,0.25,Step Function,Constant,78.172,78.53
24,0.0,0.75,Step Function,Constant,76.166,76.28
25,0.5,1.0,Step Function,Step,79.045,79.16
26,0.25,1.0,Step Function,Step,76.631,76.75
27,0.75,1.0,Step Function,Step,79.168,79.44
\end{filecontents*}
\csvreader[
            tabular={llllcc},
            table head=\toprule\bfseries Policy type & \bfseries $k_1$ & \bfseries $k_2$ & \bfseries Class & \bfseries Avg. accuracy & \bfseries Best accuracy \\\midrule,
            table foot=\bottomrule]
            {tables/data_augmentation_alpha_policy/augmentation_data.csv}
            {2=\Kone,3=\Ktwo, 4=\Policy,5=\Cls,6=\avg,7=\best}
            {\Policy & \Kone & \Ktwo & \Cls & \avg & \best}
\end{table}

% ,K1,K2,Type,Class,Avg acc,Best acc

%% file: appendix_focusing.tex
\section{Feature Focusing}

We provide details on two experiments in this section. The first involves plant classification and the second hand-gesture classification. In both cases we use \fax to focus the training on the foreground initially, and then expand by including the background. 

\subsection{Classification of abiotic stress in plants from RGB images}
\label{apendx:agrinet}

\paragraph{Background} This example deals with images of greenhouse plants treated in one of four fertilization policies.  The task required is classification of the images into one of the four treatment policies. 

\paragraph{Dataset} as mentioned in the paper, the dataset is composed of 2874 training images and 1199 test images of 51 banana plants (for each class) of various ages (plant age varies between 0-61 days). In Figure~\ref{fig:agrinet_dataset}, we show sample images from the dataset, where in the left image are the training set after \textbf{Focusing}, and in the right image are the training set after the \textbf{eXpansion}.

\paragraph{Experiment setup} the experiment setup is summarized in Table~\ref{table:expirments_summary}. Since the plant's age is part of the input, we adapt our network so that we can feed to it the plant's age as well. This is done by concatenating the features extracted by the ResNET18~\cite{he2016deep} with a one-hot vector representing the plant's age. The concatenated vector is then fed to a fully connected layer. The feature extraction part of ResNET18~\cite{he2016deep} is pre-trained on ImageNet~\cite{ImageNET}, while the fully connected layer is randomly initialized.

\paragraph{Results } in the paper, we report the top-1 accuracy of the network when trained with 4 different methods - (1) training without background using the provided segmentation, (2) training with background, (3) training with \fax-Blending implementation and (4) training with \fax-Mixing implementation. Here, we provide a visualization of where the network is looking when performing the classification. We use GradCAM~\cite{GradCAM}, a well established method for this purpose. As can be seen from Figure~\ref{fig:agrinet_grad_cam}, training with \fax yields  better localization compared with the results obtained when training with the background. This supports our claim that training with background could lead the network to look at regions that are irrelevant for the classification task - which could lead to miss-classification, or even biased models.  

% \textbf{Focus:} we start by focusing the training on the segmented images of the plant, hence ignoring other parts of the image (see left image in Figure\ref{fig:agrinet_dataset}). The idea is that the background is irrelevant for the classification, and we expect the meaningful information to be within the plant itself. \textbf{eXpand:} we gradually expand the input features by reintroducing the background in the training set using either the Mixing or Blending technique.

\input{figures/agrinet_data/different_opacities}

\input{figures/agrinet_gradcam_vis/grad_cam_vis}

\subsection{Typing-Writing Classification}
\paragraph{Background}  This example portrays images of hands typing on a virtual keyboard, that were captured at the desk level. The task required is identification of the keys pressed in the given image. Similar to the previous example, here too the background poses a distraction for the learner, whose task involves analysis of hand gestures, thus there is motivation for feature focusing.

\paragraph{Dataset} The dataset contains $613,349$ training images and $50,801$ test images, of the hands of 18 people (see Figure \ref{fig:hands_dataset}). To make the data more challenging, the hands are segmented out of their captured background (which includes the actual keyboard), and are laid over random backgrounds taken from the MS-COCO dataset~\cite{lin2014microsoft}. \textbf{Focus:} The trained model is expected to recover pressed keys by observing hand posture. Thus, the foreground, i.e.\ pixels depicting the hand within the image, are the primary focus of the model and the training process. For this reason training starts with images showing just the hands with no background. \textbf{eXpand:} As training progresses, we gradually incorporate the background into the input, making sure the model is able to handle this type of data at test time and eliminating the need for semantic segmentation in real time. With $\alpha=0$, the background is completely blacked out, and as $\alpha$ increases, the probability for incorporating a blended version of the background increases, such that for $\alpha=1$, the full background is applied. The blending parameter is drawn at random within the range $[0,\alpha]$ for any $\alpha \in (0,1)$. In Figure~\ref{fig:hands_dataset}, we provide a visualization of 100 random images from the dataset for different opacity values. 

\paragraph{Experiment setup:} The model employed here is ResNet-18~\cite{he2016deep} with an Adam optimizer, learning rate of $1\exp -4$, weight decay of $1 \exp -6$ and batch size of 32. We train the model for 35 epochs, such that the transformation of $\alpha$ occurs throughout the first 10. Thus, $\alpha$ increases by $0.1$ every epoch, and then remains unchanged when it reaches the value $1$.  The experiment setup summary appears in Table~\ref{table:expirments_summary}, with the experiment name Hands-Classification.

\input{tables/hands_classification/table}

\input{figures/hands_dataset/hands_dataset_vis}

\paragraph{Results: } In Table \ref{tbl:hands_tbl} we compare the results of \fax (bottom row) vs. three baselines. The first (No BG) does not make use of any background throughout training or testing. The second (With BG), uses the pasted-on backgrounds throughout training, and real backgrounds during testing. The third (Step), starts off without any background, and after 5 epochs, immediately adds it in. During testing, real backgrounds are used. The "No BG" baseline naturally achieves the overall best scores. However, note that \fax is a close second and allows testing on images with real backgrounds, whereas "No BG" does not. This is an important distinction, since typing classification is generally an online task that calls for speed of execution. Thus, by slightly compromising the F1 score, we are able to bypass the need to segment each testing frame prior to its classification, delivering a faster response.

%% file: figures/agrinet_data/different_opacities.tex
\begin{figure}
    \centering
    \includegraphics[width=0.18\textwidth]{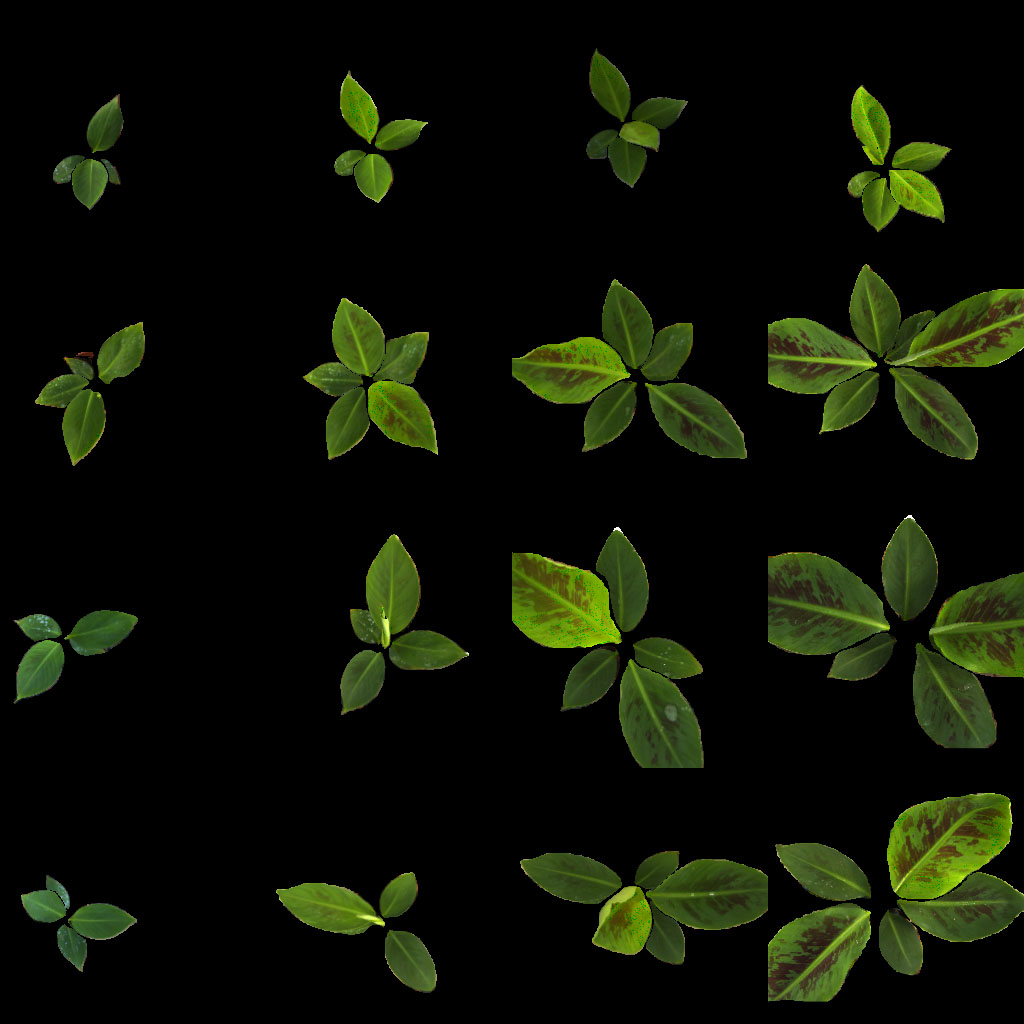}
    \includegraphics[width=0.18\textwidth]{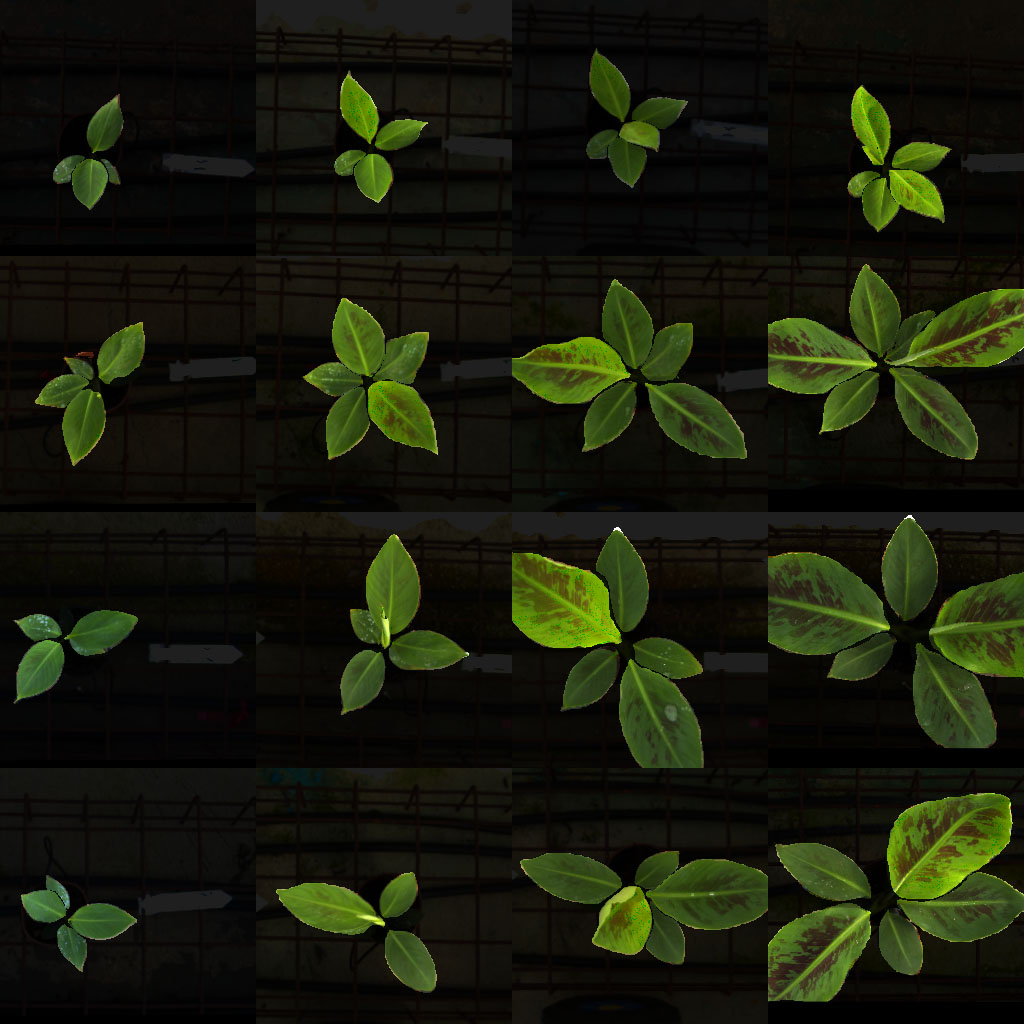}
    \includegraphics[width=0.18\textwidth]{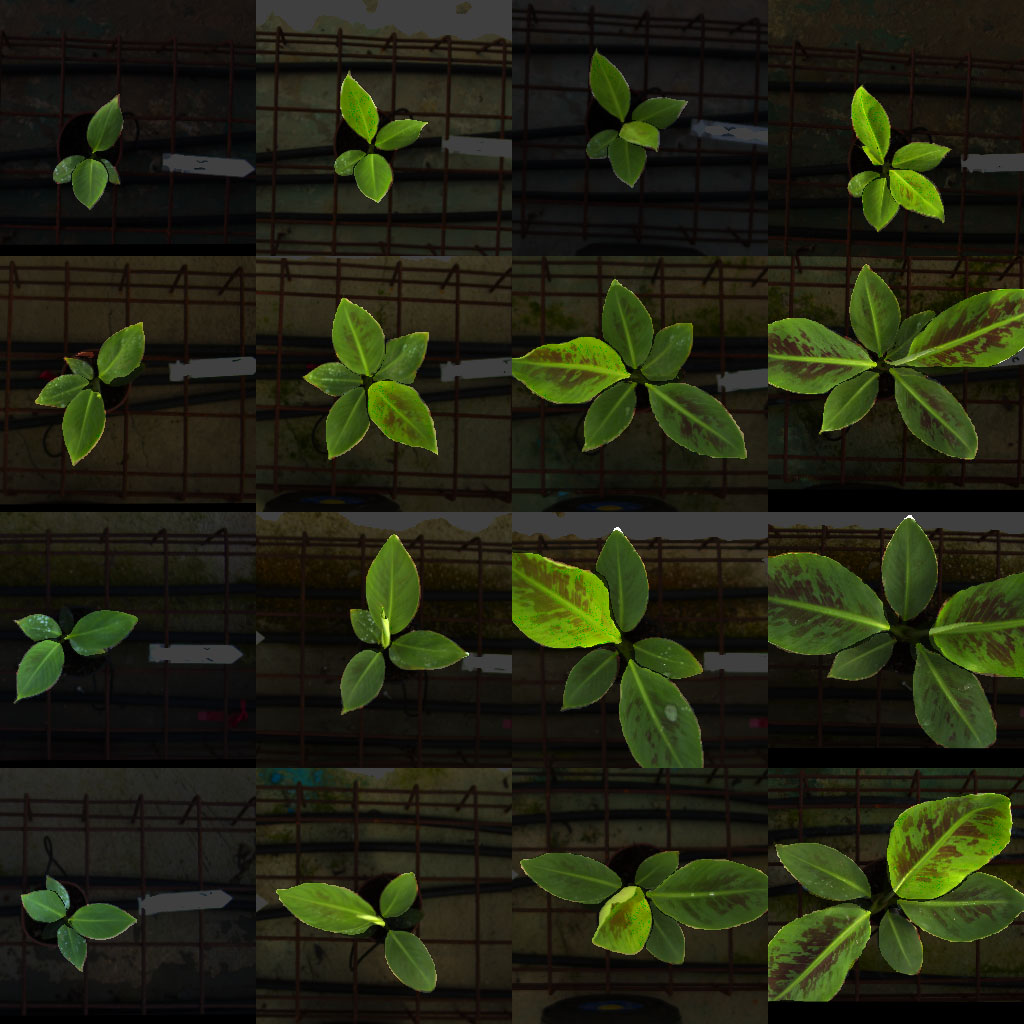}
    \includegraphics[width=0.18\textwidth]{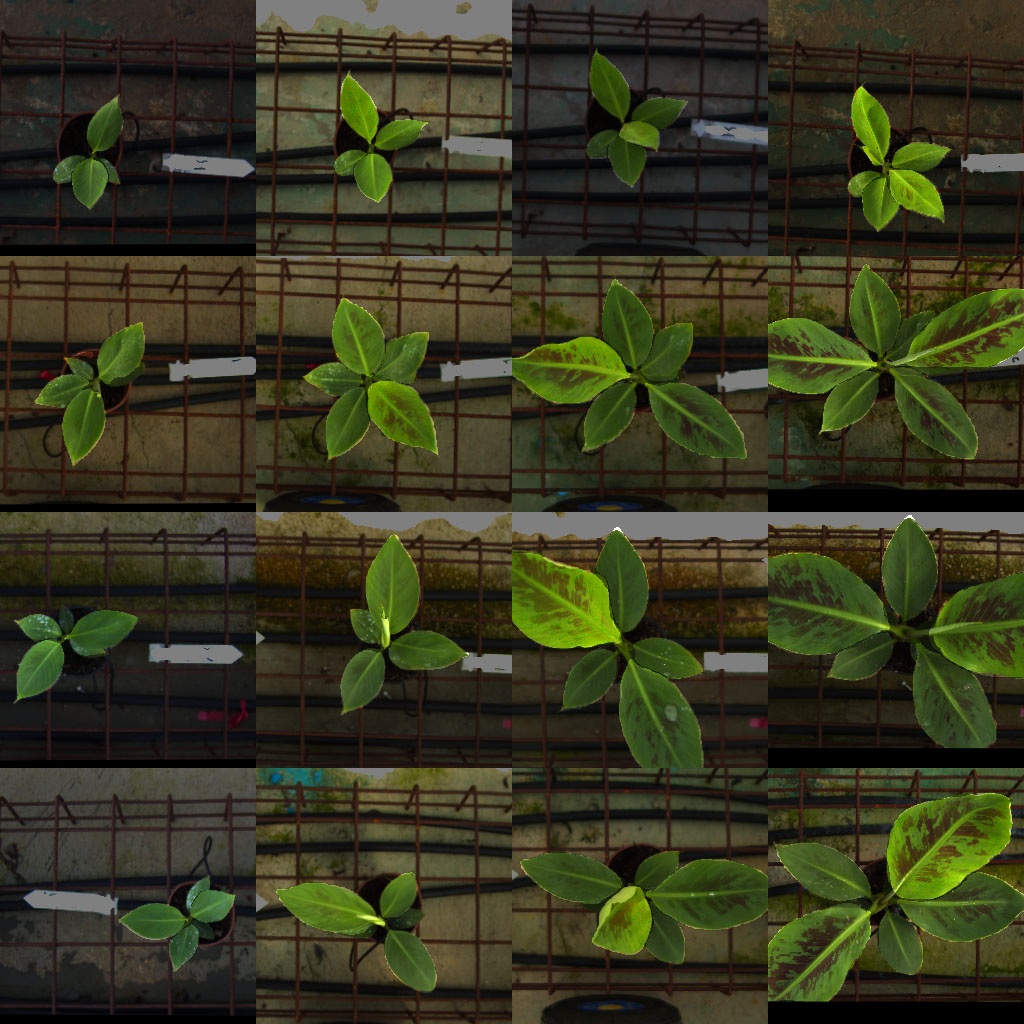}
    \includegraphics[width=0.18\textwidth]{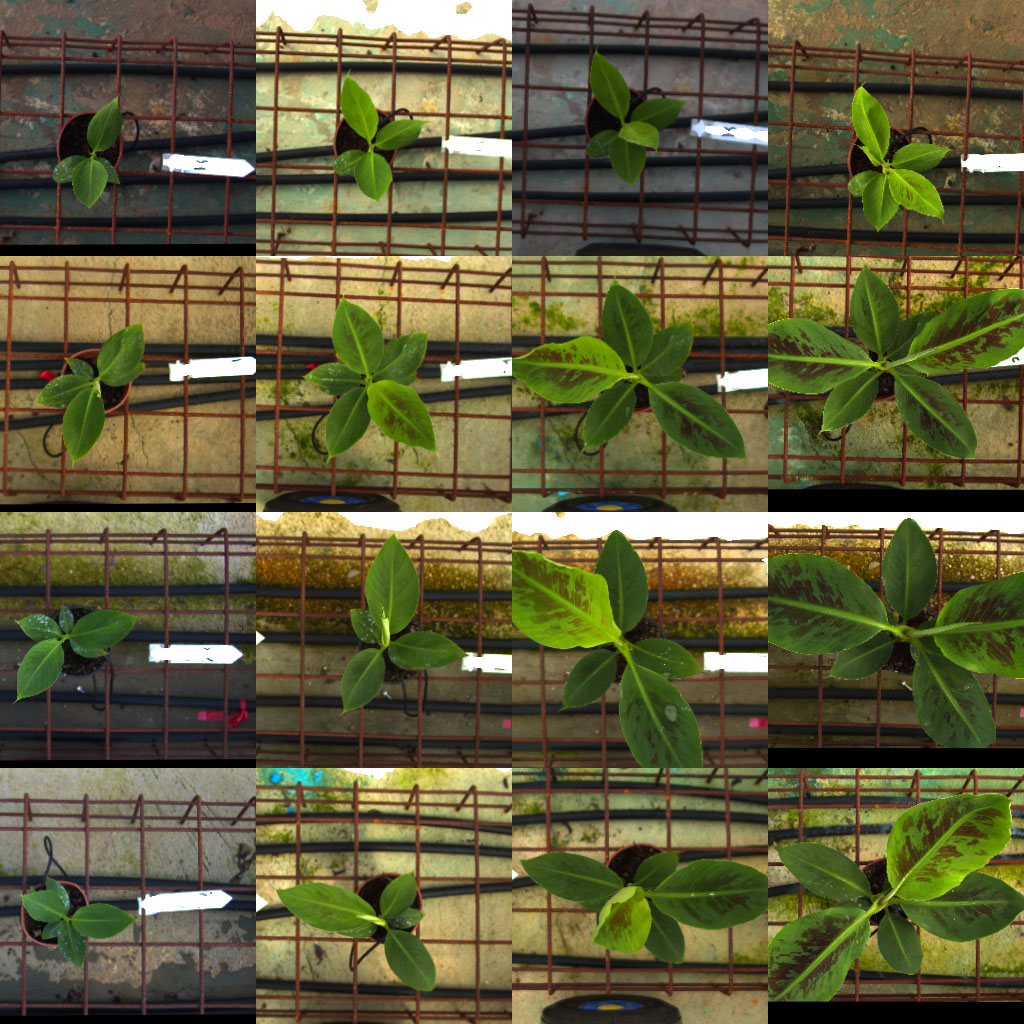}
    \caption{Visualization of the dataset used for the fertilization stress classification experiment. Each image grid (from left-to-right) visualizes the dataset with different background opacity (0\%, 12.5\%, 25\%, 50\%, 100\%). For each grid, images from different classes (rows) and different ages (columns) are presented.}
    \label{fig:agrinet_dataset}
\end{figure}

%% file: figures/agrinet_gradcam_vis/grad_cam_vis.tex
\begin{figure}
    \centering
    \begin{subfigure}{0.22\textwidth}
    \includegraphics[width=\textwidth]{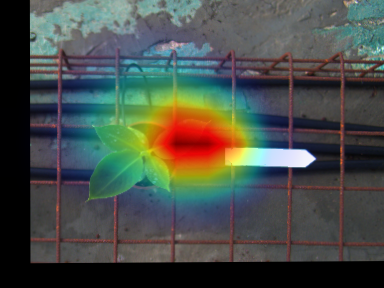}
    \includegraphics[width=\textwidth]{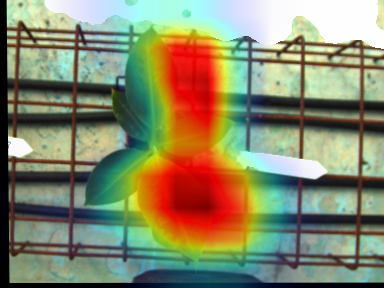}
    \includegraphics[width=\textwidth]{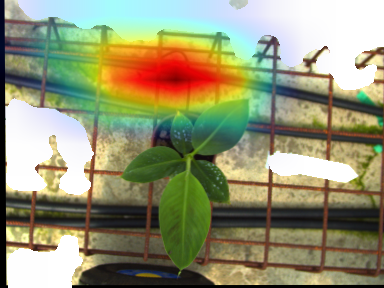}
    \includegraphics[width=\textwidth]{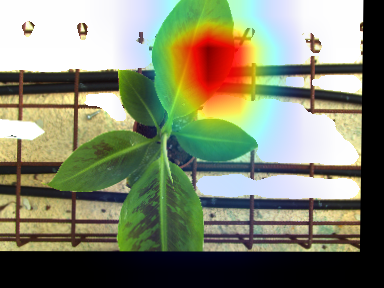}
    \caption{With BG}
    \end{subfigure}
    \begin{subfigure}{0.22\textwidth}
    \includegraphics[width=\textwidth]{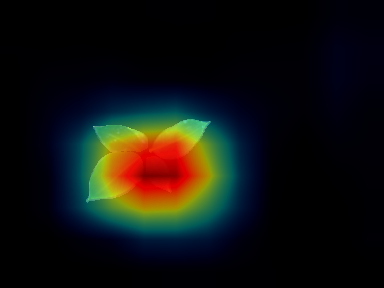}
    \includegraphics[width=\textwidth]{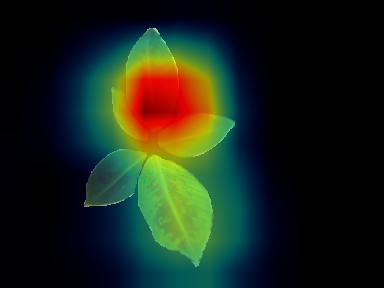}
    \includegraphics[width=\textwidth]{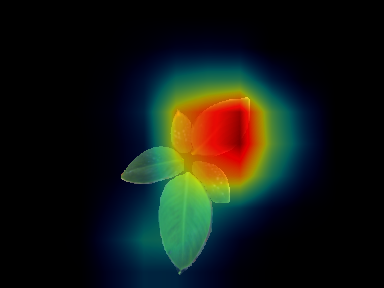}
    \includegraphics[width=\textwidth]{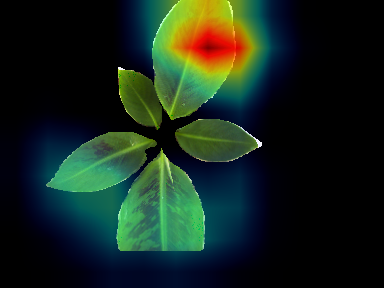}
    \caption{No BG}
    \end{subfigure}
    \begin{subfigure}{0.22\textwidth}
    \includegraphics[width=\textwidth]{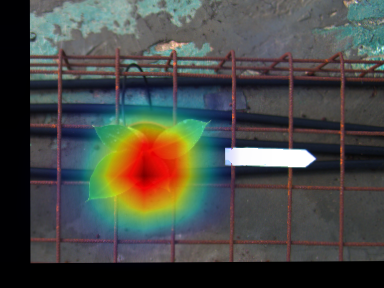}
    \includegraphics[width=\textwidth]{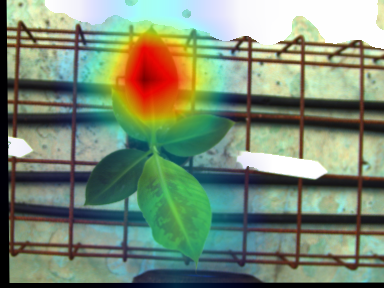}
    \includegraphics[width=\textwidth]{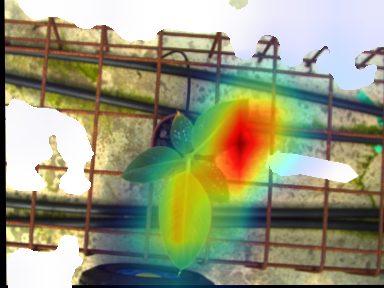}
    \includegraphics[width=\textwidth]{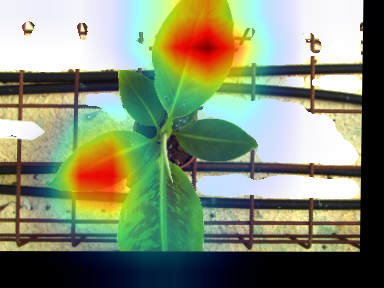}
    \caption{\fax - Blending}
    \end{subfigure}
    \begin{subfigure}{0.22\textwidth}
    \includegraphics[width=\textwidth]{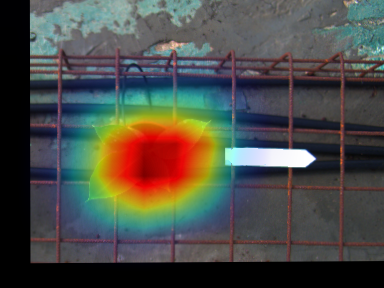}
    \includegraphics[width=\textwidth]{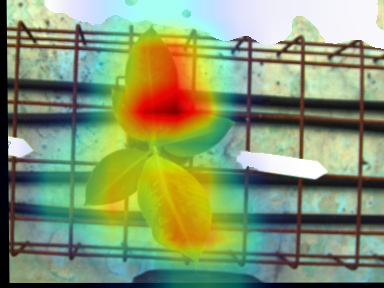}
    \includegraphics[width=\textwidth]{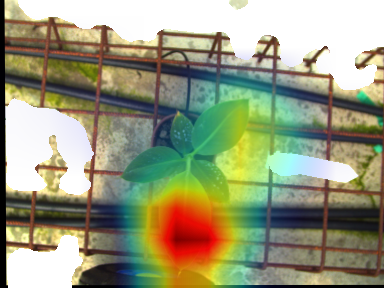}
    \includegraphics[width=\textwidth]{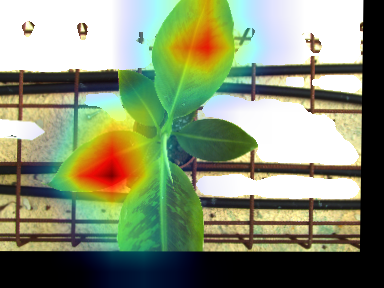}
    \caption{\fax - Mixing}
    \end{subfigure}
    \caption{GradCAM~\cite{GradCAM} visualization of four training methods -- train with background ("With BG"), train without background ("No BG"), and train using \fax (Mixing and Blending strategies). Clearly, adapting \fax obtains satisfactory results, whereas training with background may lead to noisy models. Warm (red) regions indicate the regions that mostly affected the network's decision.}
    \label{fig:agrinet_grad_cam}
\end{figure}

%% file: tables/hands_classification/table.tex
\begin{table}[ht]
    \centering
    \caption{Typing classification comparison results. Entries marked with $\dagger$ are tested on images without a background. We compare \fax (bottom row) to three baselines -- "No BG", "With BG" and "Step". See text for details and analysis.}\label{table:hands_classification}
        \begin{filecontents*}{tables/hands_classification/hands_classification.csv}
Training method,Accuracy,Recall,Precision,F1 Score
No BG$^{\dagger}$,0.824,0.701,0.825,0.758
With BG,0.750,0.588,0.791,0.675
Step,0.746,0.501,0.873,0.637
\fax,0.791,0.658,0.837,0.737
\end{filecontents*}
\csvreader[
            tabular={lcccc},
            table head=\toprule\bfseries Training method & \bfseries Accuracy& \bfseries Recall & \bfseries Precision & \bfseries F1 Score \\\midrule,
            table foot=\bottomrule]
            {tables/hands_classification/hands_classification.csv}
            {1=\method,2=\acc,3=\rec,4=\pre,5=\fone}
            {\method & \acc & \rec & \pre & \fone}
\label{tbl:hands_tbl}
\end{table}

%% file: figures/hands_dataset/hands_dataset_vis.tex
\begin{figure}
    \centering
    \includegraphics[width=0.32\textwidth]{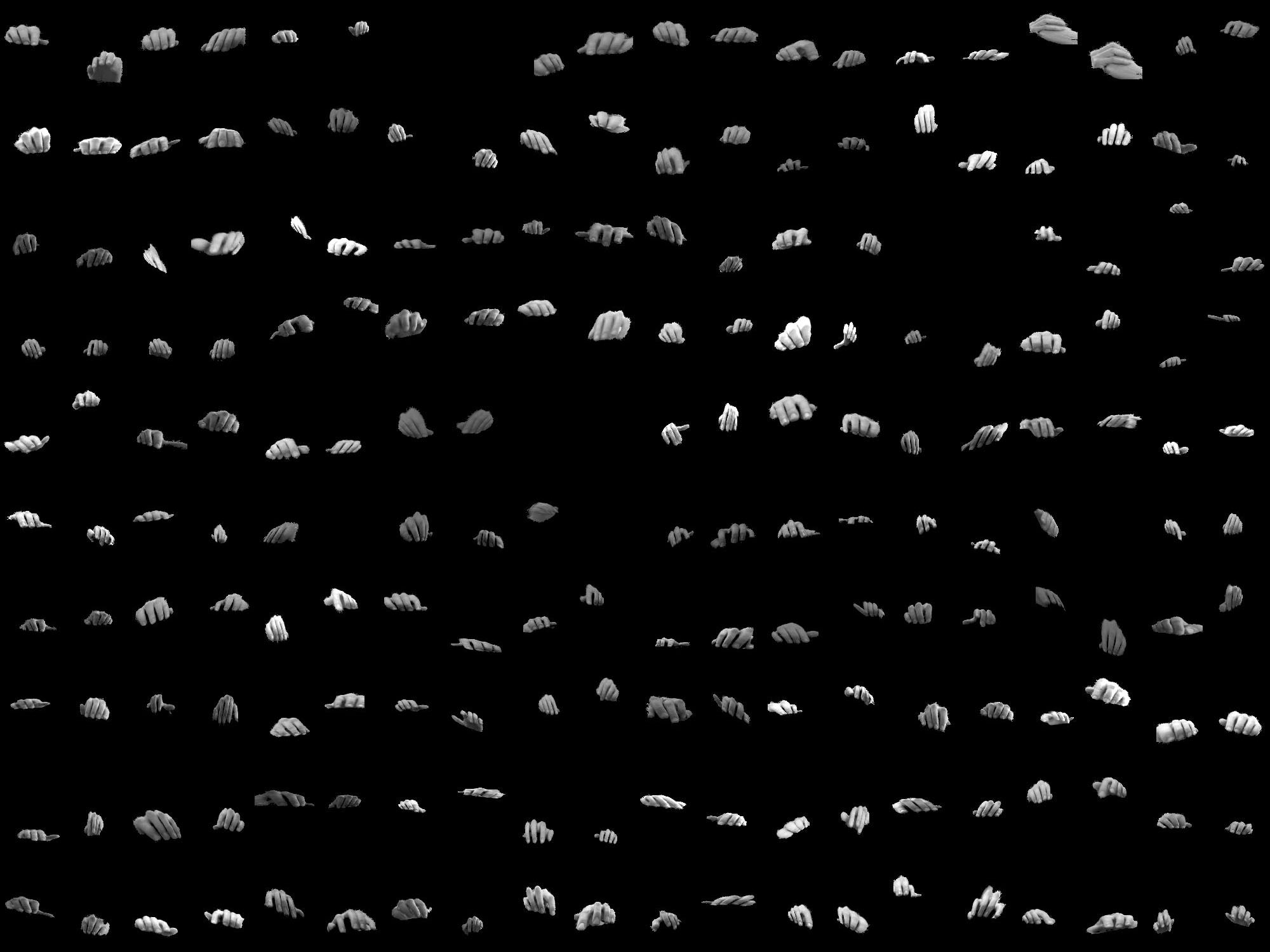}
    \includegraphics[width=0.32\textwidth]{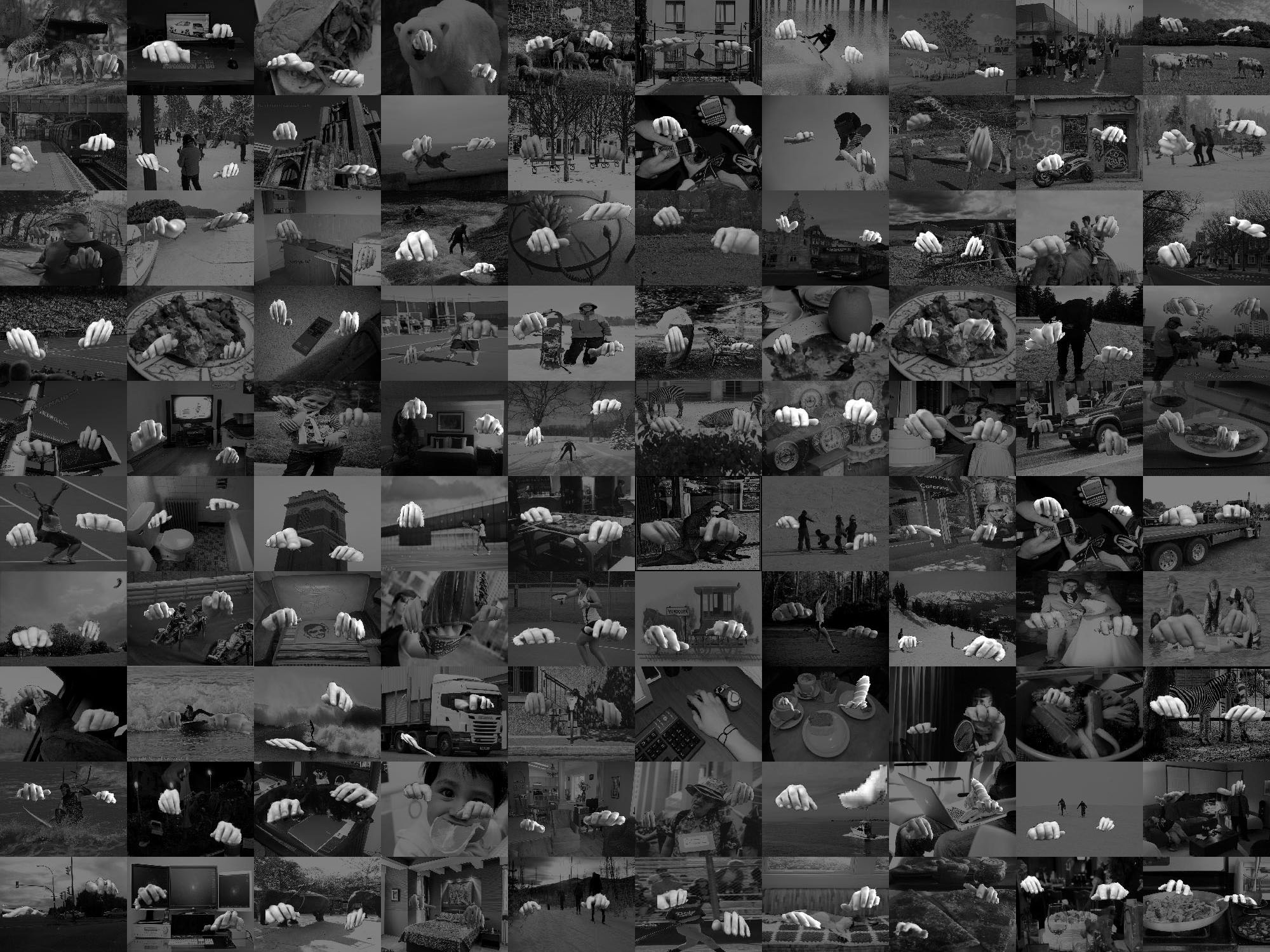}
    \includegraphics[width=0.32\textwidth]{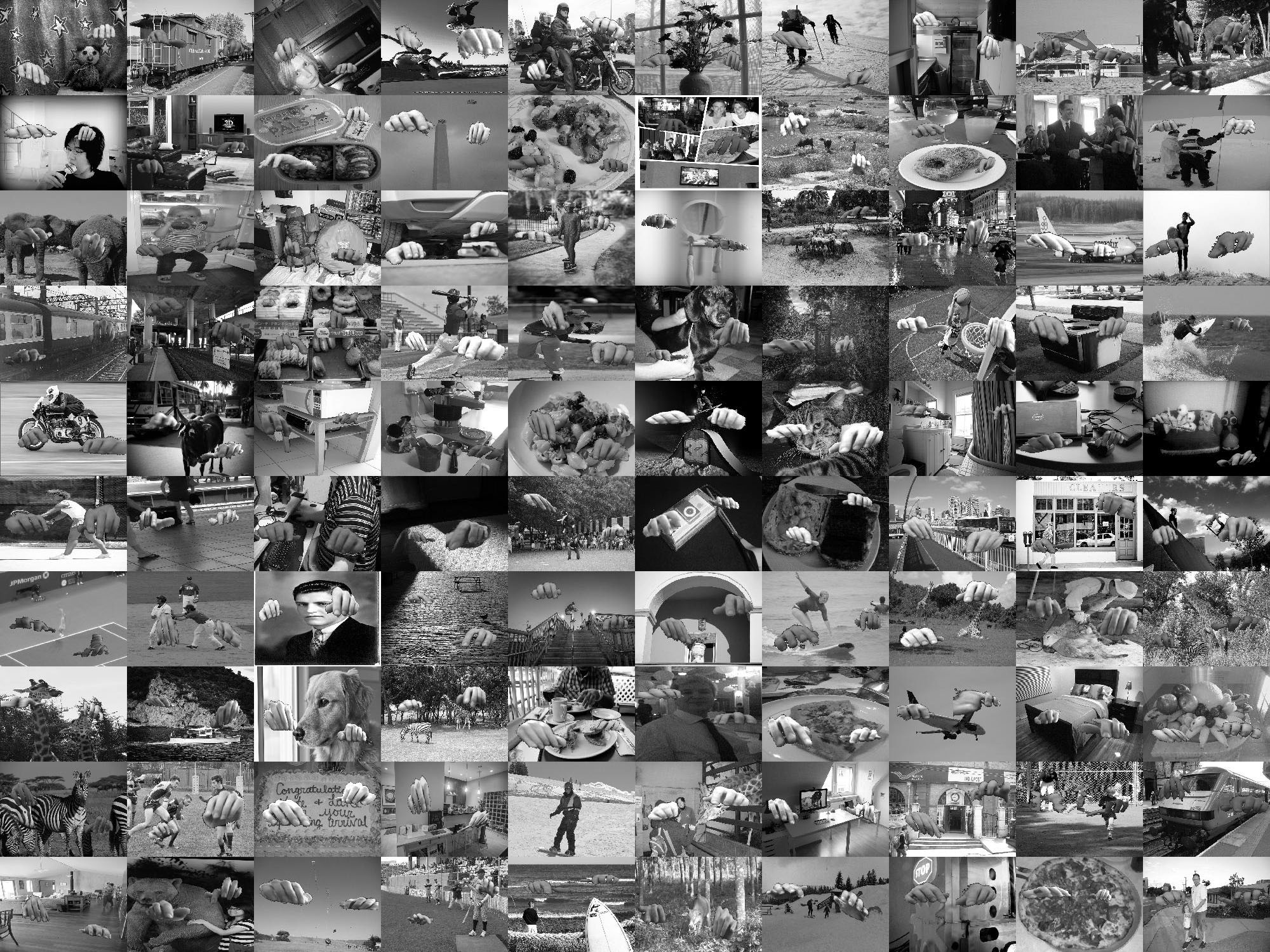}
    \caption{Typing classification dataset visualization. Left-to-right: only foreground of hands (background segmented out), original images captured, foreground pasted onto arbitrary backgrounds from MS-COCO.}
    \label{fig:hands_dataset}
\end{figure}